\documentclass[letterpaper]{article} 
\usepackage{aaai25}  
\usepackage{times}  
\usepackage{helvet}  
\usepackage{courier}  
\usepackage[hyphens]{url}  
\usepackage{graphicx} 
\usepackage{natbib}  
\usepackage{caption} 
\usepackage{algorithm}
\usepackage{algorithmic}
\usepackage{amssymb}
\usepackage{amsmath}
\usepackage{booktabs}
\usepackage{multirow}
\usepackage{array}
\newcolumntype{C}[1]{>{\centering\arraybackslash}m{#1}}

\usepackage{newfloat}
\usepackage{listings}
\DeclareCaptionStyle{ruled}{labelfont=normalfont,labelsep=colon,strut=off} 
\floatstyle{ruled}
\newfloat{listing}{tb}{lst}{}
\floatname{listing}{Listing}
\nocopyright 
\title{Dynamic Adaptive Optimization for Effective Sentiment Analysis Fine-Tuning on Large Language Models}
\author{
Hongcheng Ding\textsuperscript{\rm 1*},
Xuanze Zhao\textsuperscript{\rm 2},
Ruiting Deng\textsuperscript{\rm 3},
Shamsul Nahar Abdullah\textsuperscript{\rm 1},
Deshinta Arrova Dewi\textsuperscript{\rm 1},
Zixiao Jiang\textsuperscript{\rm 4},
}
\affiliations{
\textsuperscript{\rm 1}INTI International University, Malaysia\\
\textsuperscript{\rm 2}Zhejiang University of Finance and Economics Dongfang College, China\\
\textsuperscript{\rm 3}Honghe Autonomous Prefecture 3th Hospital, China\\
\textsuperscript{\rm 4}Chinese Academy of Medical Sciences and Peking Union Medical College, China\\
i24025877@student.newinti.edu.my
}
\begin{document}
\maketitle

\begin{abstract}
Sentiment analysis plays a crucial role in various domains, such as business intelligence and financial forecasting. Large language models (LLMs) have become a popular paradigm for sentiment analysis, leveraging multi-task learning to address specific tasks concurrently. However, LLMs with fine-tuning for sentiment analysis often underperforms due to the inherent challenges in managing diverse task complexities. Moreover, constant-weight approaches in multi-task learning struggle to adapt to variations in data characteristics, further complicating model effectiveness. To address these issues, we propose a novel multi-task learning framework with a dynamic adaptive optimization (DAO) module. This module is designed as a plug-and-play component that can be seamlessly integrated into existing models, providing an effective and flexible solution for multi-task learning. The key component of the DAO module is dynamic adaptive loss, which dynamically adjusts the weights assigned to different tasks based on their relative importance and data characteristics during training. Sentiment analyses on a standard and customized financial text dataset demonstrate that the proposed framework achieves superior performance. Specifically, this work improves the Mean Squared Error (MSE) and Accuracy (ACC) by 15.58\% and 1.24\% respectively, compared with previous work.
\end{abstract}

\section{Introduction}
Sentiment analysis has become an essential tool for businesses to understand and analyze customer opinions and feedback, gauging public perception of their products, services, and brand reputation \cite{liu2020sentiment,alaei2019sentiment}. It is also used in demand forecasting by analyzing online product reviews \cite{kharfan2021data,li2022using} and in supply chain management to monitor supplier performance and identify potential risks \cite{sharma2020covid}. In the financial sector, sentiment analysis is crucial for forecasting stock market trends \cite{bai2023financial}, predicting cryptocurrency prices \cite{chowdhury2020approach}, and forecasting exchange rates \cite{ding2024eurusd}.

\begin{figure}
    \centering
    \includegraphics[width=1.0\linewidth]{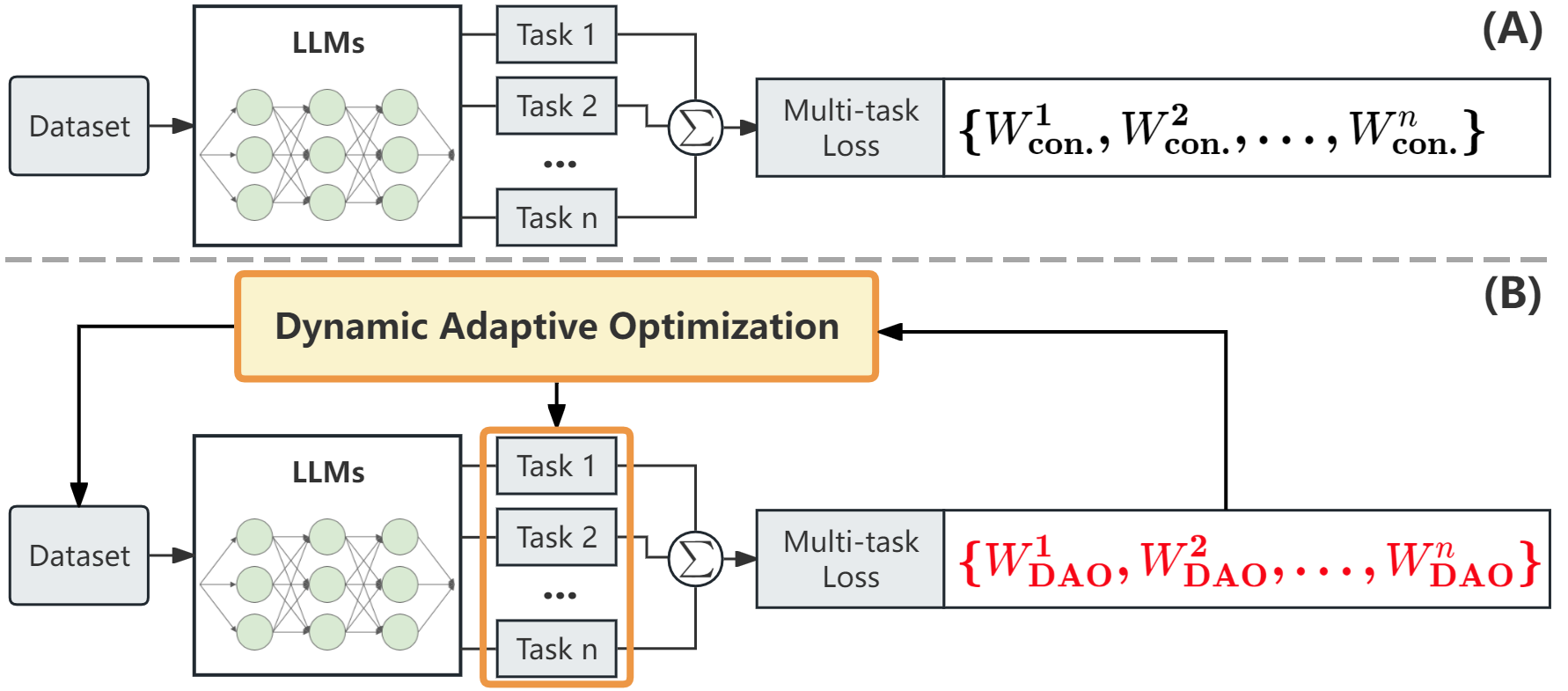}
    \caption{Illustration of the traditional multi-task learning approach with constant weights \textbf{(A)} and our proposed multi-task learning framework with the DAO module \textbf{(B)}. In \textbf{(A)}, constant weights $\{W^{1}_{\text{con.}}, W^{2}_{\text{con.}}, \dots, W^{n}_{\text{con.}}\}$ are used to combine the losses of Task 1 to Task $n$. In \textbf{(B)}, dynamic adaptive loss adjusts the weights $\{W^{1}_{\text{DAO}}, W^{2}_{\text{DAO}}, \dots, W^{n}_{\text{DAO}}\}$ dynamically during the training process based on the current batch of data and the losses of different tasks.}
    \label{fig:main plot}
\end{figure}

Traditionally, sentiment analysis employs various approaches, primarily including machine learning algorithms (e.g., SVM and Naive Bayes \cite{dang2020sentiment}) and deep learning methods (e.g., CNN and LSTM \cite{zhang2018deep}). Recently, Large Language Models (LLMs) such as BERT \cite{devlin2018bert}, GPT \cite{radford2018improving}, and RoBERTa \cite{liu2019roberta} have revolutionized Natural Language Processing (NLP) by achieving state-of-the-arts performance across various NLP tasks. For instance, \citeauthor{hande2021benchmarking} (\citeyear{hande2021benchmarking}) explore multi-task learning for sentiment analysis and offensive language identification, while \citeauthor{ding2024boosting} (\citeyear{ding2024boosting}) conduct aspect-based sentiment analysis.

However, the conventional approach of fine-tuning these models by assigning constant weights to the loss functions of different tasks in the overall loss function does not always guarantee optimal convergence. This limitation is particularly pronounced when dealing with non-standard datasets or in scenarios with limited data availability \cite{kendall2018multi}. Consequently, a notable performance gap often emerges between the fine-tuned model and the anticipated performance under ideal conditions.

To address these challenges, we propose a multi-task learning framework with the dynamic adaptive optimization (DAO) module. Specifically, by incorporating multi-task learning \cite{houlsby2019parameter}, the model can capture more nuanced sentiment patterns even with limited data, facilitating more effective convergence. Additionally, the introduction of the DAO module allows the model to dynamically adjust the weights of different loss functions in each training iteration (see Figure \ref{fig:main plot}). The key component of the DAO module is dynamic adaptive loss, which enables the model to adapt to the evolving learning process and optimize the balance between different tasks, ultimately leading to improved overall performance. The main contributions of this paper are as follows:

\begin{itemize}
\item We identify a performance gap in LLMs fine-tuned for sentiment analysis within the financial domain using multi-task learning, particularly when employing constant-loss weights for sub-tasks. 

\item We design a plug-and-play DAO module that effectively resolves the problem in multi-task learning. The module dynamically adjusts the weights of different tasks based on the relevance and importance of each data batch and the varying difficulty of tasks.

\item To reduce the computational overhead of fine-tuning the model, we use LoRA with different parameters to fine-tune the RoBERTa-Large model for the sentiment analysis task without the introduction of inference latency.

\item Our framework achieves the best performance, improving Mean Squared Error (MSE) by 15.58\% and Accuracy (ACC) by 1.24\% compared to previous works.

\end{itemize}

\section{Motivation}

In the sentiment polarity analysis task for exchange rate texts, we observe that the performance of the RoBERTa-Large \cite{liu2019roberta} is unsatisfactory after fine-tuning on the standard and customized financial text dataset \cite{ding2024eurusd}. This can be attributed to the model's lack of exposure to news domain-specific texts containing specialized jargon, implicit sentiments, and subtle variations. Subsequently, we employ an  alternative RoBERTa-Large model, Twitter-RoBERTa-Large \cite{loureiro2023tweet} fine-tuned on a tweet news dataset, and the model's performance shows a slight improvement.

\begin{figure}[h]
\centering
\includegraphics[width=1\columnwidth]{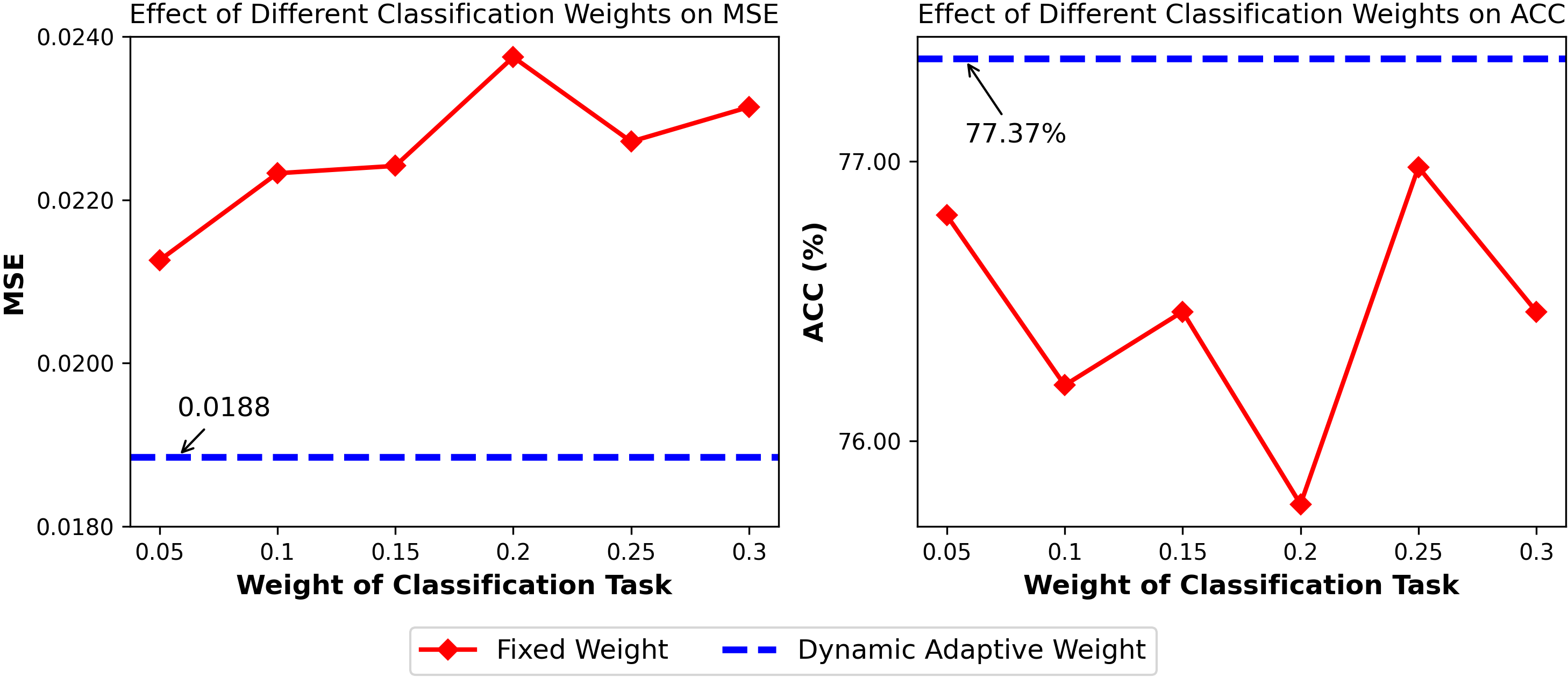}
\caption{Effect of different classification task weights on MSE and ACC in multi-task learning, compared with the performance of our proposed framework incorporating the DAO module. 
}
\label{fig:weight performance}
\end{figure}

In the subsequent multi-task learning experiments, we modulate the weight ($W_c$) of the classification task within the total loss function, with the regression task weight correspondingly set to ($1-W_c$). Figure \ref{fig:weight performance} illustrates the model’s task performance across different tasks as a result of varying weights. As the weight of the classification task increases, the MSE increases while the ACC decreases, indicating a trade-off between the regression and classification objectives. The optimal balance appears to be around a classification weight of 0.25, where both MSE is relatively low and ACC is relatively high. Upon integration of the DAO, the model achieves superior task performance across all metrics, yielding the lowest MSE of 0.0188 and the highest ACC of 77.37\%. 

\section{Method}
In this section, we will introduce traditional sentiment analysis, our framework incorporating the DAO module with LoRA.
\subsection{Traditional Sentiment Analysis}
\begin{figure*}[t]
\centering
\includegraphics[width=1\textwidth]{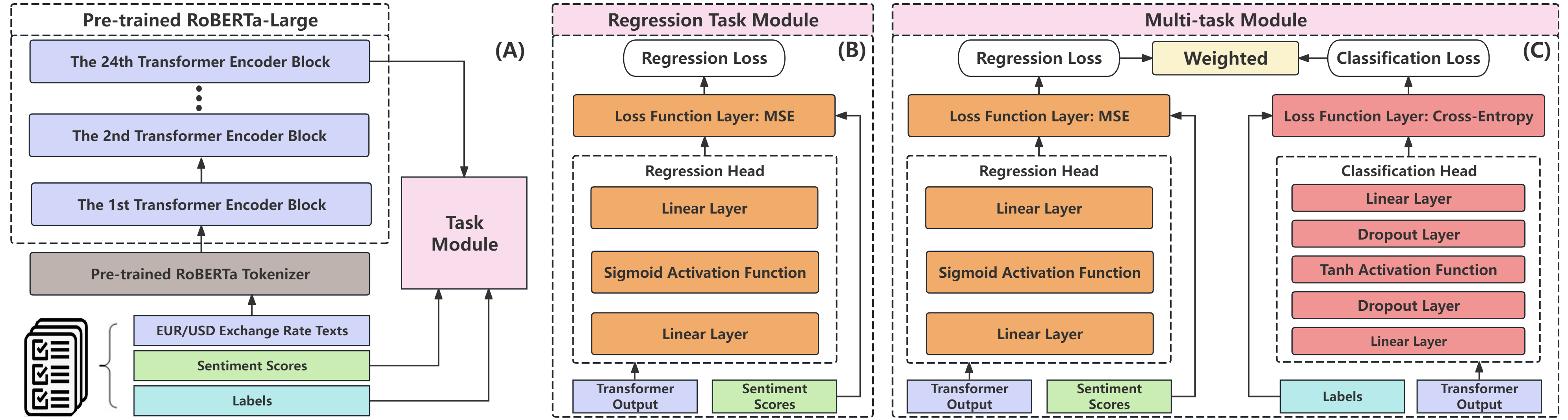}
\caption{The diagram outlines the application of the pre-trained RoBERTa-Large model for sentiment analysis, segmented into three main parts: (A) RoBERTa-Large as the backbone combined with a task-specific module, (B) a regression task module, and (C) a multi-task module that handles both regression and classification tasks.}
\label{fig:model architecutre}
\end{figure*}

\noindent\textbf{Backbone: RoBERTa-Large.}
In sentiment analysis, LLMs serve as text data embedding generators, concatenating various task-specific heads for various sentiment analysis tasks. This work employs RoBERTa-Large as the embedding generator for text data, as illustrated in Figure \ref{fig:model architecutre} (A). The training set is defined as \( D = \{(x_i, y_i)\}_{i=1}^N \), where \( N \) denotes the total number of texts, \( x_i \) represents each text in the dataset, and \( y_i \in [-1, 1] \) corresponds to the annotated sentiment polarity score. 

The pre-trained RoBERTa tokenizer \( \mathrm{Tokenizer}(\cdot) \) processes each text \( x_i \) in the dataset, and then the tokenized text passes through 24 Transformer encoder blocks $\text{Encoder}(\cdot)$ to generate the final hidden state $H_i$:
\begin{equation}
H_i = \mathrm{Encoder}( \mathrm{Tokenizer}(x_i)),
\end{equation}
where \( H_i \in \mathbb{R}^{1 \times 1024} \). This hidden state serves as input to task-specific heads. For sentiment polarity analysis, we implement a regression head (Figure \ref{fig:model architecutre} (B)) comprising two linear layers and a sigmoid activation function:
\begin{equation}
S_i = \mathrm{LL}_2(\sigma(\mathrm{LL}_1(H_i))),
\end{equation}
where $\mathrm{LL}_1: \mathbb{R}^{1024} \rightarrow \mathbb{R}^{128}$ is the first linear layer, $\sigma(\cdot)$ denotes the sigmoid function, and $\mathrm{LL}_2: \mathbb{R}^{128} \rightarrow \mathbb{R}$ produces the final sentiment polarity score $S_i \in [-1, 1]$.

For the regression task, we use Mean Squared Error (MSE) as the loss function, denoted as \( L_{\mathrm{r}} \):
\begin{equation}
L_{\mathrm{r}} = \frac{1}{n} \sum_{i=1}^{n} (\hat{y}_i - y_i)^2
\end{equation}
where $n$ denotes the number of texts texts in the training set, $\hat{y}_i$ represents the predicted polarity score, and $y_i$ is the ground truth sentiment polarity score.

\noindent\textbf{Standard Learning.}
In sentiment analysis, particularly in data-limited scenarios, enhancing model performance is crucial. We introduce a multi-task learning approach. Specifically, we incorporate a classification task alongside the regression task to better capture the sentiment tendencies embedded in the text. This approach mitigates the impact of noise and small fluctuations while effectively alleviating the data sparsity problem often encountered in sentiment analysis. \cite{saif2012alleviating,kim2013sentiment}

Based on empirical observations, we map the continuous sentiment polarity scores to discrete sentiment categories ranging from 0 to 4, representing strong negative, negative, neutral, positive, and strong positive sentiments, respectively. The mapping function is defined as:
\begin{equation}
z_i = \begin{cases} 
4, & \text{if } y_i > 0.5; \\ 
3, & \text{if } 0.049 < y_i \leq 0.5; \\ 
2, & \text{if } -0.049 \leq y_i \leq 0.049; \\ 
1, & \text{if } -0.5 \leq y_i < -0.049; \\ 
0, & \text{if } y_i < -0.5.
\end{cases}
\end{equation}
Here, $z_i$ denotes the discrete sentiment classification of $x_i$. Consequently, we redefine our dataset as $D = \{(x_i, y_i, z_i)\}_{i=1}^N$.

For the classification task, we implement a classification head as illustrated in Figure \ref{fig:model architecutre} (C). This head comprises two dropout layers, two linear layers, and a Tanh activation function. The process can be formalized as follows:
\begin{equation}
H_i' = \mathrm{LL}_1(\mathrm{DP}_1(H_i)),
\end{equation}
where $H_i' \in \mathbb{R}^{1 \times 128}$ is an intermediate representation, $\mathrm{DP}_1(\cdot)$ is the first dropout layer for regularization, and $\mathrm{LL}_1: \mathbb{R}^{1024} \rightarrow \mathbb{R}^{128}$ is the first linear layer.

The classification label $C_i$ for the $i$-th text is then computed as:
\begin{equation}
C_i = \mathrm{LL}_2(\mathrm{DP}_2(\mathrm{Tanh}(H_i'))),
\end{equation}
where $\mathrm{Tanh}(\cdot)$ introduces non-linearity, $\mathrm{DP}_2(\cdot)$ is the second dropout layer, and $\mathrm{LL}_2: \mathbb{R}^{128} \rightarrow \mathbb{R}^5$ maps to the five discrete classification labels.

For the classification task, we use Cross-Entropy (CE) loss to measure the difference between predicted probabilities and actual class labels. The classification loss \( L_{\mathrm{c}} \) is formulated as:
\begin{equation}
L_{\mathrm{c}} = -\sum_{i=1}^{n} y_i \log(\hat{y}_i)
\end{equation}
where \( n \) is the number of texts. \( y_i \) is the actual class label, and \( \hat{y}_i \) is the predicted class label. 

To jointly optimize the regression and classification tasks, we define a multi-task learning loss $L_{\mathrm{mtl}}$:
\begin{equation}
\label{eq:mtt_loss}
L_{\mathrm{mtl}} = w_r L_r + w_c L_c,
\end{equation}
where $L_r$ and $L_c$ are the regression and classification losses, respectively, and $w_r$ and $w_c$ are hyperparameters that manage the trade-off between the tasks.

\subsubsection{Stratified Sampling Algorithm}

Let $\mathcal{Y}$ be a continuous space representing sentiment polarity scores, where each score $y \in \mathcal{Y}$ is a real number bounded by $[-1, 1]$. Our objective is to map each $y \in \mathcal{Y}$ to a discrete sentiment category $z$ in a finite set $\mathcal{Z} = \{z_1, z_2, \dots, z_k\}$, where $k$ is the total number of sentiment categories. We define a mapping function $f: \mathcal{Y} \rightarrow \mathcal{Z}$ that assigns each $y \in \mathcal{Y}$ to a corresponding $z \in \mathcal{Z}$ based on a set of predefined thresholds $\mathcal{T} = \{\tau_0, \tau_1, \dots, \tau_{k-1}\}$, where $-1 < \tau_0 < \tau_1 < \dots < \tau_{k-1} < 1$. The stratified sampling algorithm can be formally described as follows:

Given:
\begin{itemize}
    \item Continuous sentiment polarity score space $\mathcal{Y} \subseteq [-1, 1]$
    \item Discrete sentiment category set $\mathcal{Z} = \{z_1, z_2, \dots, z_k\}$
    \item Threshold set $\mathcal{T} = \{\tau_0, \tau_1, \dots, \tau_{k-1}\}$, where $-1 < \tau_0 < \tau_1 < \dots < \tau_{k-1} < 1$
\end{itemize}

Mapping Function:
\[
f: \mathcal{Y} \rightarrow \mathcal{Z}, \text{ where for each } y \in \mathcal{Y},
\]
\[
f(y) = 
\begin{cases}
z_1, & \text{if } y < \tau_0 \\
z_2, & \text{if } \tau_0 \leq y < \tau_1 \\
\vdots \\
z_k, & \text{if } y \geq \tau_{k-1}
\end{cases}
\]
The stratified sampling algorithm maps each continuous sentiment polarity score $y \in \mathcal{Y}$ to a discrete sentiment category $z \in \mathcal{Z}$ based on the predefined thresholds in $\mathcal{T}$. This process discretizes the continuous sentiment space, enabling the application of classification techniques in sentiment analysis.

\subsubsection{Sentiment Category Distribution}

Let $\mathcal{D} = \{(x_1, y_1), (x_2, y_2), \dots, (x_n, y_n)\}$ be a dataset of size $n$, where $x_i$ represents a text sample and $y_i \in \mathcal{Y}$ is its corresponding sentiment polarity score. After applying the stratified sampling algorithm, we obtain a new dataset $\mathcal{D}' = \{(x_1, y_1, z_1), (x_2, y_2, z_2), \dots, (x_n, y_n, z_n)\}$, where $z_i = f(y_i) \in \mathcal{Z}$ is the discrete sentiment category assigned to $y_i$.

To analyze the distribution of sentiment categories in $\mathcal{D}'$, we define a probability mass function $p: \mathcal{Z} \rightarrow [0, 1]$ for each sentiment category $z \in \mathcal{Z}$:
\[
p(z) = \frac{|\{i: z_i = z\}|}{n},
\]
where $|\{i: z_i = z\}|$ denotes the number of samples in $\mathcal{D}'$ with sentiment category $z$, and $n$ is the total number of samples.

The probability mass function $p(z)$ represents the distribution of sentiment categories in the dataset $\mathcal{D}'$. By analyzing this distribution, we can gain insights into the prevalence and balance of different sentiment levels in the data, informing further data preprocessing or model design decisions.

The resulting dataset $\mathcal{D}' = \{(x_i, y_i, z_i)\}_{i=1}^n$ obtained from the stratified sampling algorithm and sentiment category distribution analysis contains rich information for sentiment analysis tasks. The original continuous sentiment polarity scores $y_i$ are preserved, enabling regression tasks, while the discrete sentiment categories $z_i$ facilitate classification tasks. This dataset supports multi-task learning approaches that leverage both continuous and discrete sentiment information to improve sentiment analysis performance.

\subsection{Dynamic Adaptive Optimization}

In the context of multi-task learning with LLMs, two primary challenges arise: \textit{Task-level challenge: inter-task difficulty discrepancy.} The loss functions associated with regression and classification tasks may exhibit significant disparities in their magnitudes. Consequently, this discrepancy can lead to the dominance of one task over others during the training process, potentially hindering the model's ability to effectively learn and generalize across all tasks. \textit{Data-level challenge: imbalanced data distribution.} Within each training batch, the sample distribution can exhibit considerable variability, and the class distribution may be inherently imbalanced. As a result, the model may be susceptible to overfitting or underfitting for certain classes, compromising its overall performance and generalization capability.

\begin{figure}[t]
\centering
\includegraphics[width=1\columnwidth]{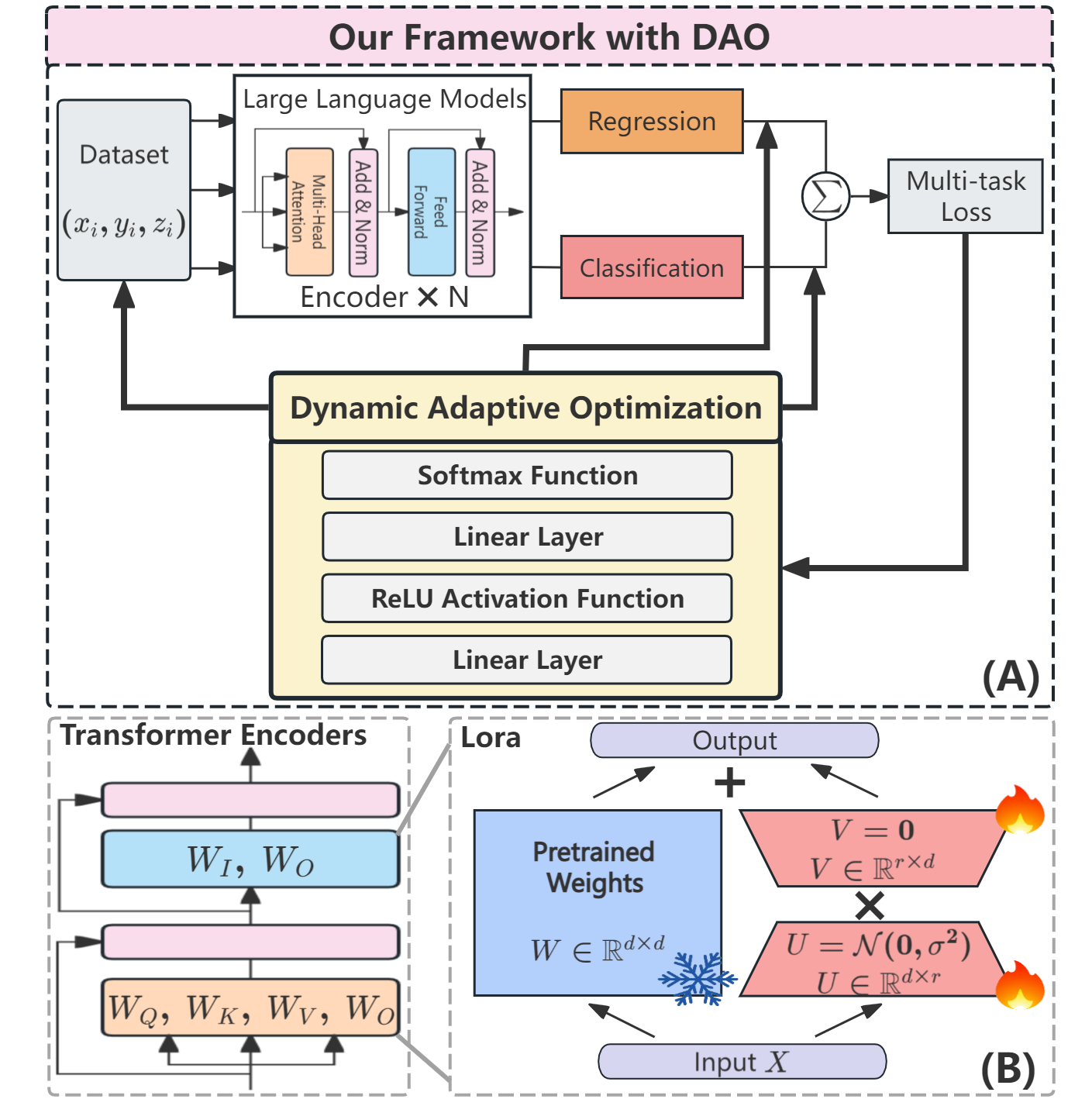}
\caption{(A) Workflow of the Multi-Task Learning Module with DAO, and (B) Application of LoRA Technique}
\label{fig:lora}
\end{figure}

To address these challenges, we propose a DAO module as shown in Figure \ref{fig:lora} (A), which is a learnable neural network integrated into the multi-task learning framework. 

\noindent\textbf{Solution \#1: Inter-task Difficulty Discrepancy.} Due to the different magnitudes of the loss functions for regression and classification tasks, directly summing them may cause one task to dominate the entire training process. To balance the contributions of different tasks, we introduce a gradient-based weighting method. Specifically, we define the total multi-task loss function \(L^t_{\text{mtl}}\) as follows:
\begin{equation}
L^t_{\text{mtl}} = \lambda_r^t w_r^t L_r^t + \lambda_c^t w_c^t L_c^t,
\end{equation}
where \(L_r^t\) and \(L_c^t\) represent the losses of the regression and classification tasks at step \(t\), respectively, \(w_r^t\) and \(w_c^t\) are the task-specific weights, and \(\lambda_r^t\) and \(\lambda_c^t\) are the gradient-based weighting coefficients. \\
At each training step, we compute the gradients of the regression and classification task losses:
\begin{equation}
g_r^t = \nabla L_r^t, \quad
g_c^t = \nabla L_c^t.
\end{equation}
Then, we use the L2 norms of these gradients to compute the gradient-based weighting coefficients:
\begin{equation}
\lambda_r^t = \frac{|g_c^t|}{|g_r^t| + |g_c^t|}, \quad
\lambda_c^t = \frac{|g_r^t|}{|g_r^t| + |g_c^t|}.
\end{equation}
These gradient-based weighting coefficients are used to balance the learning progress of different tasks and scale them to similar magnitudes in the total multi-task loss function \(L^t_{\text{mtl}}\).

\noindent\textbf{Solution \#2: Imbalanced Data Distribution.}
We introduce a regularization term and class-specific weights into the classification loss function. The regularization term \(-\alpha \log p_k^t\) encourages the model to pay more attention to minority classes with smaller sample proportions, while the class-specific weights \(v_k^t\) are inversely proportional to the sample proportions \(p_k^t\):
\begin{equation}
v_k^t = \frac{1}{(p_k^t)^\beta},
\end{equation}
where \(\beta\) is a parameter that controls the degree of class balancing.

We use \(D\) to denote the entire dataset and \(D^t\) to represent the batch of data sampled at time step \(t\). For each batch \(D^t\), we calculate the proportion of samples \(p_k^t\) belonging to class \(k\) as:
\begin{equation}
p_k^t = \frac{|D_k^t|}{|D^t|},
\end{equation}
where \(|D_k^t|\) and \(|D^t|\) denote the number of samples in batch \(D^t\) belonging to class \(k\) and the total number of samples in batch \(D^t\), respectively.

To incorporate the regularization term \(-\alpha \log p_k^t\) and the class-specific weights \(v_k^t\) into the classification loss function, we define the imbalanced data distribution loss \(L_{\text{imb}}^t\) conditioned on \(D^t\) as:
\begin{equation}
L_{\text{imb}}^t(D^t) = \sum_{k=1}^B v_k^t (p_k^t L_{c,k}^t - \alpha \log p_k^t).
\end{equation}
This loss captures the sample distribution and class proportions specific to the current batch, allowing the model to dynamically adapt to the characteristics of each batch during training. We then incorporate the imbalanced data distribution loss \(L_{\text{imb}}^t(D^t)\) into the total multi-task loss function \(L^t_{\text{mtl}}\) from Challenge 1, along with the task-specific weights \(w_r^t\) and \(w_c^t\):
\begin{align}
L_{\text{mtl}}^t(D^t) &= \lambda_r^t w_r^t L_r^t + \lambda_c^t w_c^t L_{\text{imb}}^t(D^t) \nonumber \\
&= \lambda_r^t w_r^t L_r^t + \lambda_c^t w_c^t \sum_{k=1}^B v_k^t (p_k^t L_{c,k}^t - \alpha \log p_k^t),
\end{align}
where \(\lambda_r^t\) and \(\lambda_c^t\) are the gradient-based weighting coefficients, \(v_k^t\) are the class-specific weights, and \(L_{c,k}^t\) is the classification loss for class \(k\) at time step \(t\). The total multi-task loss \(L_{\text{mtl}}^t\) is now conditioned on the current batch \(D^t\), allowing the model to adapt to the specific characteristics of each batch during training.

To learn the hyperparameters \(\alpha\) and \(\beta\), we treat them as learnable parameters of the DAO Module and update them using gradient descent along with the other parameters of the module. During the backward pass, the gradients of the DAO Module loss \(L_{\text{mtl}}^t(D^t)\) with respect to \(\alpha\) and \(\beta\) are computed as follows:
\begin{equation}
\frac{\partial L_{\text{mtl}}^t(D^t)}{\partial \alpha} = -\lambda_c^t w_c^t \sum_{k=1}^B v_k^t \log p_k^t.
\end{equation}
To compute the gradient with respect to \(\beta\), we apply the chain rule and substitute the expression for \(\frac{\partial v_k^t}{\partial \beta}\):
\begin{align}
\frac{\partial L_{\text{mtl}}^t(D^t)}{\partial \beta} &= -\lambda_c^t w_c^t \sum_{k=1}^B \frac{\partial v_k^t}{\partial \beta} (p_k^t L_{c,k}^t - \alpha \log p_k^t) \nonumber \\
&= \lambda_c^t w_c^t \sum_{k=1}^B \frac{\log p_k^t}{(p_k^t)^\beta} (p_k^t L_{c,k}^t - \alpha \log p_k^t).
\end{align}
The hyperparameters \(\alpha\) and \(\beta\) are then updated using an optimizer such as Adam or AdamW.

To obtain the task-specific weights \(w_r^t\) and \(w_c^t\), we pass the gradient-weighted regression loss \(\lambda_r^t L_r^t\), the gradient-weighted imbalanced data distribution loss \(\lambda_c^t L_{\text{imb}}^t(D^t)\), and the current batch \(D^t\) through the DAO Module:
\begin{equation}
w_{r}^{t}, w_{c}^{t} = \text{DAO}(\lambda_{r}^{t} L_{r}^{t}, \lambda_{c}^{t} L_{\text{imb}}^{t}(D^t), D^t)
\end{equation}

Specifically, we first concatenate these weighted losses and map them to a hidden space through a fully connected layer (FC$_1$) followed by a ReLU activation function:
\begin{align}
\mathbf{h}^t &= \text{ReLU}(\text{FC}_1([\lambda_r^t L_r^t, \lambda_c^t L_{\text{imb}}^t(D^t)])) \nonumber \\
&= \text{ReLU}(\text{FC}_1([\lambda_r^t L_r^t, \lambda_c^t \sum_{k=1}^B v_k^t (p_k^t L_{c,k}^t - \alpha \log p_k^t)])).
\end{align}
The hidden representation \(\mathbf{h}^t\) encodes the information from the weighted losses and the current batch data. Next, we use another fully connected layer (FC$_2$) to map the hidden representation \(\mathbf{h}^t\) to a two-dimensional vector \(\mathbf{s}^t\):
\begin{equation}
\mathbf{s}^t = \text{FC}_2(\mathbf{h}^t).
\end{equation}
Finally, we pass the vector \(\mathbf{s}^t\) through a Softmax function to obtain the task-specific weights \(w_r^t\) and \(w_c^t\) based on the current batch \(D^t\):
\begin{equation}
[w_r^t, w_c^t] = \text{Softmax}(\mathbf{s}^t).
\end{equation}

In summary, the DAO Module receives batch-level information including losses, gradients, sample distributions, and the current batch \(D^t\). The DAO module dynamically adjusts the weights, allowing the model to adapt to the characteristics of different batches, balance the learning progress of tasks, and mitigate the data imbalance problem. The gradient-based weighting coefficients \(\lambda_r^t\) and \(\lambda_c^t\) help address the inter-task difficulty discrepancy, while the learnable hyperparameters \(\alpha\) and \(\beta\) are directly related to the imbalanced data distribution and are optimized using gradient descent to minimize the multi-task learning loss \(L_{\text{mtl}}^t\).

\subsection{LoRA}

We implement LoRA \cite{hu2021lora} for parameter-efficient fine-tuning (PEFT) of the pre-trained RoBERTa-large model. LoRA injects trainable low-rank decomposition matrices into each layer, reducing parameters while maintaining performance.

LoRA introduces rank-$r$ matrices $U \in \mathbb{R}^{d \times r}$ and $V \in \mathbb{R}^{r \times d}$ into each attention block and feed-forward network, as shown in Figure \ref{fig:lora} (B). The figure illustrates how LoRA is integrated into the Transformer encoder, with pretrained weights $W$ remaining frozen while low-rank matrices are trained. The adjusted weight matrix $W'$ during forward propagation is:
\begin{equation}
f(x) = (W + UV) \cdot X + b,
\end{equation}
where $b$ is the bias, and $X$ is the input as illustrated in Figure \ref{fig:lora} (B). Only $U$ and $V$ are updated during training, significantly reducing trainable parameters and potentially accelerating training.

In traditional multi-task learning with constant weight, this approach can lead to suboptimal performance. Our proposed framework incorporates the DAO module that dynamically adjusts task weights based on their relative importance, gradients, and data characteristics during training. This batch-level dynamic adaptive loss addresses the limitations of constant-weight approaches by considering the varying difficulties and contributions of each task and the characteristics of the data in each batch. The plug-and-play nature of the DAO module allows for flexible integration into various multi-task learning scenarios, making it applicable to a wide range of tasks.

\section{Evaluation}

\subsection{Software and Hardware } 
We use Ubuntu 22.04, Python 3.9.19, PyTorch 2.3.1, PEFT 0.12.0, and CUDA 12.4, running on a system with 32GB RAM and an NVIDIA RTX 3090Ti GPU with 24GB VRAM.

\subsection{Setup and Dataset}
\noindent{\textbf{Benchmark.}} We use diffrent benchmarks to verify our framework. We establish three benchmark: RoBERTa-Large \cite{liu2019roberta} for regression, Twitter-RoBERTa-Large \cite{loureiro2023tweet} for regression, and Twitter-RoBERTa-Large with constant-weight multi-task learning for regression and classification. Additionally, we propose two methods: Twitter-RoBERTa-Large with DAO for multi-task learning, and the multi-task learning framework using Twitter-RoBERTa-Large with DAO and LoRA.

\begin{table*}[!htb]
\centering
\resizebox{0.915\linewidth}{!} {
\begin{tabular}{|C{1.0cm}|C{1.5cm}|C{1.5cm}|C{1.5cm}|C{1.5cm}|C{1.4cm}
                         |C{1.5cm}|C{1.5cm}|C{1.5cm}|C{1.5cm}|C{1.4cm}|}
\hline
\multirow{2}{*}{Epoch} & \multicolumn{5}{c|}{RoBERTa-Large for Regression} & \multicolumn{5}{c|}{Twitter-RoBERTa-Large for Regression} \\
\cline{2-11}
 & MSE & MAE & RMSE & \(R^2\) (\%) & GPU & MSE & MAE & RMSE & \(R^2\) (\%) & GPU \\
\hline
20 & 0.019064 & 0.098594 & 0.138074 & 78.59 & 21,780 & 0.021369 & 0.106370 & 0.146183 & 76.00 & 21,780 \\
40 & 0.019805 & 0.099945 & 0.140731 & 77.76 & MiB & 0.019398 & 0.100493 & 0.139277 & 78.21 & MiB \\
\cline{6-6} \cline{11-11}
60 & 0.018328 & 0.096009 & 0.135719 & 78.86 & Time & 0.019353 & 0.099428 & 0.139116 & 78.27 & Time \\
\cline{6-6} \cline{11-11}
80 & 0.018299 & 0.095898 & 0.137581 & 78.74 & 130,525 & 0.018617 & 0.097000 & 0.136446 & 79.09 & 130,553 \\
100 & 0.018949 & 0.095773 & 0.137657 & 78.72 & seconds & 0.018714 & 0.097313 & 0.136801 & 78.98 & seconds \\
\hline
\end{tabular}
}
\caption{Performance Metrics of Single Task Experiments.}
\label{table_b1_b2}
\end{table*}

\begin{table*}[!htb]
\centering
\resizebox{0.915\linewidth}{!} {
\begin{tabular}{|C{1.0cm}|C{1.5cm}|C{1.5cm}|C{1.5cm}|C{1.5cm}|C{1.4cm}
                         |C{1.5cm}|C{1.5cm}|C{1.5cm}|C{1.5cm}|C{1.4cm}|}
\hline
\multirow{2}{*}{Epoch} & \multicolumn{5}{c|}{Multi-Task Learning with constant Weight} & \multicolumn{5}{c|}{Multi-Task Learning with DAO} \\
\cline{2-11}
 & MSE & ACC & Precision & F1 & GPU & MSE & ACC & Precision & F1 & GPU \\
\hline
20 & 0.022544 & 74.83 & 76.37 & 75.39 & 21,780 & 0.018983 & 75.60 & 76.99 & 75.99 & 22,336 \\
40 & 0.022930 & 75.90 & 76.39 & 76.01 & MiB & 0.018668 & 76.64 & 77.92 & 76.83 & MiB \\
\cline{6-6} \cline{11-11}
60 & 0.023744 & 75.52 & 75.94 & 75.61 & Time & 0.018841 & 76.68 & 78.06 & 76.94 & Time \\
\cline{6-6} \cline{11-11}
80 & 0.022612 & 76.08 & 75.97 & 75.88 & 129,959 & 0.019000 & 76.89 & 77.98 & 77.00 & 211,024 \\
100 & 0.022326 & 76.11 & 76.17 & 76.05 & seconds & 0.018848 & 77.06 & 78.28 & 77.37 & seconds \\
\hline
\end{tabular}
}
\caption{Performance Metrics of Multi-Task Learning Experiments.}
\label{table_b3_b4}
\end{table*}

\begin{table*}[!htb]
\centering
\resizebox{0.915\linewidth}{!} {
\begin{tabular}{|C{1.0cm}|C{1.5cm}|C{1.5cm}|C{1.5cm}|C{1.5cm}|C{1.4cm}
                         |C{1.5cm}|C{1.5cm}|C{1.5cm}|C{1.5cm}|C{1.4cm}|}
\hline
\multirow{2}{*}{Epoch} & \multicolumn{5}{c|}{LoRA (Rank = 128)} & \multicolumn{5}{c|}{LoRA (Rank = 256)} \\
\cline{2-11}
 & MSE & ACC & Precision & F1 & Resource & MSE & ACC & Precision & F1 & Resource \\
\hline
20 & 0.020855 & 75.13 & 76.73 & 75.56 & 22,852 & 0.020905 & 75.13 & 77.44 & 75.78 & 24,224 \\
40 & 0.020442 & 75.13 & 77.25 & 75.80 &  MiB & 0.019155 & 76.72 & 78.46 & 77.00 & MiB \\
60 & 0.019320 & 76.55 & 78.10 & 76.93 & --------- & 0.019064 & 76.42 & 78.39 & 76.87 & --------- \\
80 & 0.019026 & 76.25 & 77.93 & 76.65 & 247,452 & 0.018804 & 75.82 & 77.84 & 76.25 & 266,674 \\
100 & 0.019057 & 76.33 & 78.02 & 76.76 & seconds & 0.018747 & 76.03 & 78.03 & 76.47 & seconds \\
\hline

Epoch & \multicolumn{5}{c|}{LoRA (Rank = 384)} & \multicolumn{5}{c|}{LoRA (Rank = 512)} \\
\cline{2-11}

\hline
20 & 0.020939 & 75.17 & 76.82 & 75.76 & 19,550 & 0.020835 & 75.00 & 76.68 & 75.66 & 22,336 \\
40 & 0.019411 & 75.09 & 76.94 & 75.52 & MiB & 0.018324 & 76.85 & 78.27 & 76.97 & MiB \\
60 & 0.018673 & 75.73 & 77.52 & 76.14 & --------- & 0.018376 & 76.12 & 77.74 & 76.31 & --------- \\
80 & 0.018651 & 75.39 & 77.03 & 75.69 & 286,793 & 0.018580 & 76.76 & 77.67 & 76.65 & 324,013 \\
100 & 0.018610 & 75.47 & 77.04 & 75.68 & seconds & 0.018571 & 76.38 & 77.40 & 76.35 & seconds \\
\hline
\end{tabular}
}
\caption{Performance Metrics of LoRA at Different Ranks.}
\label{tab:performance_metrics_top_lora}
\end{table*}

\noindent{\textbf{Evaluation Metrics.}} For the regression task, we utilize MSE, Mean Absolute Error (MAE), Root Mean Squared Error (RMSE), and the Coefficient of Determination (\(R^2\)). For the classification task, we use ACC, weighted Precision, and weighted F1 score.

\noindent{\textbf{Dataset.}}
We use a standard and customized financial text dataset that comprises 23,242 preprocessed news and analysis texts on EUR/USD exchange rates \cite{ding2024eurusd}. This dataset is randomly split into training and validation sets with a 9:1 ratio, and all experiments utilize these two identical datasets. Further information and supplementary details can be found in the Appendix.

\section{Main Results}
In this section, we will first introduce the hyperparameters and then analyze the experimental results.

\noindent{\textbf{Model Hyperparameters.}} The training texts exceeding 512 tokens are truncated to the first 512 tokens. The model is trained with a batch size of 10 for 100 epochs, using the AdamW optimizer with a learning rate of 1e-5 and an epsilon value of 1e-8. The training process incorporates 1e2 warmup steps and implements cosine decay for learning rate scheduling. To ensure the reproducibility of the experiments, we set constant random seeds for Python (42), NumPy (42), and PyTorch (42) across all runs.

\noindent{\textbf{DAO Module Hyperparameters.}} The optimizer is Adam with a learning rate of 0.001.

\noindent{\textbf{LoRA Parameters.}} Different LoRA configurations are applied in the experiments, with the rank values set to 8, 16, 32, 64, 128, 256, 384, and 512. Correspondingly, the scaling factor (alpha) for the LoRA matrices is set to one times the rank. The dropout rate for all LoRA configurations is constant at 0.05 to prevent overfitting.

\noindent{\textbf{Results.}} Table \ref{table_b1_b2} presents the performance metrics of RoBERTa-Large and Twitter-RoBERTa-Large on the task of sentiment polarity regression. After 100 epochs of training, RoBERTa-Large achieves an MSE of 0.018949, MAE of 0.095773, RMSE of 0.137657, and \( R^2 \) of 78.72\%, while Twitter-RoBERTa-Large demonstrates a 1.24\% improvement in MSE, a 1.61\% increase in MAE, a 0.62\% improvement in RMSE, and a 0.26\% increase in \( R^2 \).

These results indicate that Twitter-RoBERTa-Large outperforms RoBERTa-Large, likely due to its fine-tuning on a tweet news dataset, which is more similar to the target domain of financial news compared to the general domain pre-training of RoBERTa-Large. Although there is an improvement in performance, there is still a gap between the current results and the ideal performance. Therefore, a new framework may be necessary to better understand and analyze sentiment in the financial domain.

Table \ref{table_b3_b4} compares the performance of Twitter-RoBERTa-Large using multi-task learning with a constant weight of 0.1 for the classification task and 0.9 for the regression task, and our proposed framework with the DAO module. After 100 epochs of training, the proposed method achieves an MSE of 0.018848, an ACC of 77.06\%, a Precision of 78.28\%, and an F1 score of 77.37\%. Compared to the constant-weight method, the DAO module yields a 15.58\% improvement in MSE, a 1.25\% improvement in ACC, a 2.77\% improvement in Precision, and a 1.74\% improvement in F1 score.
\begin{figure}[!htb]
\centering
\includegraphics[width=1\columnwidth]{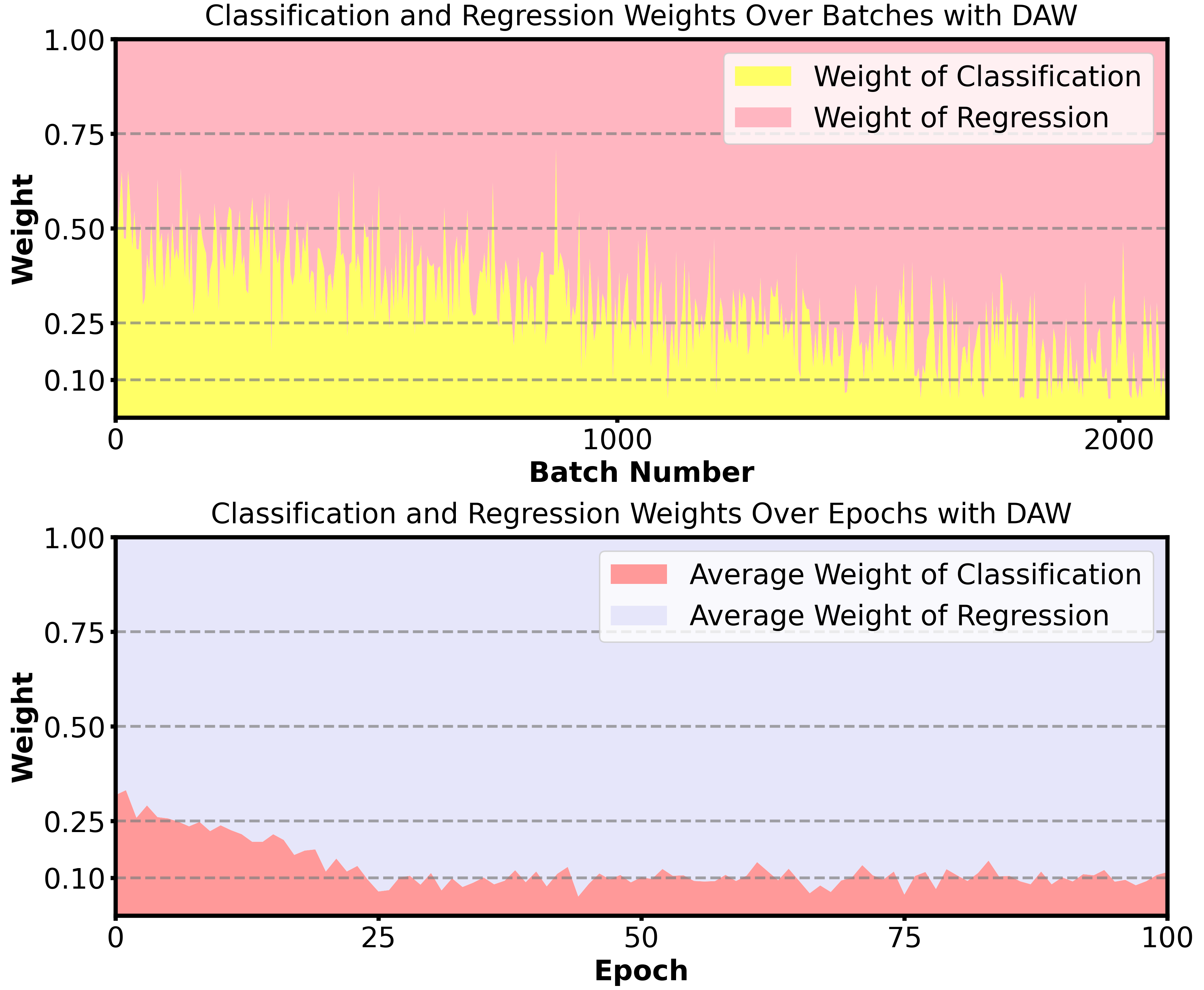}
\caption{Dynamic adjustment of task weights by the DAO module.}
\label{fig:Dynamic Adaptive Weight}
\end{figure}

Figure \ref{fig:Dynamic Adaptive Weight} illustrates the changes in the dynamic adaptive loss function during model training, with the weights of the classification and regression tasks set to 1. The upper plot shows the dynamic adjustment of task weights during training. The initial weight of the classification task appears at approximately 0.5 and exhibits an oscillating decrease, stabilizing around 0.15 after approximately 1600 batches. The lower plot displays the average weights of the classification and regression tasks throughout the entire training process. The average weight of the classification task gradually decreases from about 0.3 to 0.1 over the first 25 epochs and continues to fluctuate around 0.1 for the remainder of the training.

Additionally, significant differences in the loss function values for the two tasks are observed throughout the training process, with the discrepancy consistently remaining within at least one order of magnitude. When using the DAO module, the loss function accelerates the convergence of model training. Compared to the constant-weight loss function with a weight of 0.1, the loss function value significantly decreases within the first 50 epochs (see subsection Multi-task Loss in Appendix).

These observations demonstrate the DAO module's effectiveness in adjusting task weights based on relative difficulty and gradients, assigning higher weight to the more challenging regression task and lower weight to the easier classification task. This dynamic adjustment balances the learning progress of both tasks, optimizing overall performance, unlike constant-weight methods that may fail to capture changes in task importance and data characteristics.

\section{Ablation Studies}
We ablate our method with DAO using different LoRA ranks, as shown in Table \ref{tab:performance_metrics_top_lora}, which presents several intriguing observations. Increasing the rank consistently decreases MSE while improving ACC, Precision, and F1 score, suggesting that more trainable parameters enhance model performance. Table \ref{tab:performance_metrics_best_lora} illustrates that at a rank of 256, we achieve results comparable to a fully fine-tuned (FF) model with only 80.88\% of the FF time. Using the same amount of time, MSE is slightly improved. At a rank of 384, the optimal performance is achieved in 53.00\% of the FF time. When the rank increases to 512, we obtain results comparable to the FF model with only 36.85\% of the FF time, and the optimal performance is obtained in 66.02\% of the time, with MSE improving by 3.68\% and ACC improving by 0.12\%. Due to CUDA out of memory issues, the batch size was set to 7 in the experiment with a rank of 384, and to 10 in the experiment with a rank of 512.

\begin{table}[!h]
\centering
\resizebox{0.85\linewidth}{!} {
\begin{tabular}{@{}cccccc@{}}
\toprule
Rank & Epoch & MSE & ACC & Time & Usage \\
\midrule
256 & 64 & 0.018838 & 76.16 & 170,671 s & 80.88\% \\
256 & 79 & 0.018665 & 75.82 & 210,672 s & 99.83\% \\
384 & 39 & 0.018434 & 75.69 & 111,849 s & 53.00\% \\
512 & 23 & 0.018730 & 76.38 & 77,763 s & 36.85\% \\
512 & 42 & 0.018154 & 77.15 & 139,326 s & 66.02\% \\
\bottomrule
\end{tabular}
}
\caption{Best Performance Metrics of LoRA at Different Ranks.}
\label{tab:performance_metrics_best_lora}
\end{table}

The tables in the LoRA subsection of the Appendix illustrate the performance of LoRA fine-tuning across various ranks. The data indicate that higher ranks contribute to enhanced model performance, demonstrating that an increase in rank positively influences outcomes. The corresponding line charts reveal that as the number of epochs increases, both MSE and ACC initially decrease, reaching optimal performance, before subsequently rising again. This pattern suggests that while the model's generalization ability improves with training, the limited parameter count leads to a deterioration in metrics beyond a certain number of epochs, indicative of overfitting.

These results underscore the effectiveness of LoRA in reducing fine-tuning computational costs while maintaining performance, with a trade-off between enhanced performance and increased parameter count and training time at higher ranks.

\section{Related Work}

\noindent{\textbf{LLMs}} NLP advancements lead to the widespread application of LLMs in sentiment analysis tasks. ChatGPT shows significant potential in automating student feedback analysis, outperforming traditional deep learning models \cite{shaikh2023exploring}. Pre-trained models like BERT achieve state-of-the-arts results by learning contextual word representations \cite{liao2021improved}. In Arabic sentiment analysis, transformer-based models like RoBERTa and XLNet push boundaries despite language complexities \cite{alduailej2022araxlnet}. \citeauthor{krugmann2024sentiment} (\citeyear{krugmann2024sentiment}) reveals that GPT-3.5, GPT-4, and Llama 2 can compete with and sometimes surpass traditional transfer learning methods in sentiment analysis. \citeauthor{carneros2023comparative} (\citeyear{carneros2023comparative}) highlights the versatility of pre-trained LLMs like GPT-3.5 in diverse NLP applications, including emotion recognition. However, fully fine-tuning these LLMs for specific tasks remains computationally expensive and time-consuming, posing challenges for both academia and industry.

\noindent{\textbf{PEFT}} To address computational challenges, researchers introduce PEFT methods for LLMs. \citeauthor{hu2023llm} (\citeyear{hu2023llm}) presents LLM-Adapters, integrating adapters into LLMs and achieving comparable performance to powerful 175B parameter models using only 7B parameters in zero-shot tasks. \citeauthor{lei2023conditiona} (\citeyear{lei2023conditiona}) introduces Conditional Adapters (CODA), which adds sparse activation and new parameters to pre-trained models for efficient knowledge transfer, significantly speeding up inference. \citeauthor{hu2021lora} (\citeyear{hu2021lora}) proposes LoRA, which injects trainable low-rank decomposition matrices into each Transformer layer, reducing parameters while maintaining performance on par with full fine-tuning.

\section{Conclusion}
In this work, we propose a multi-task learning framework with a DAO module for LLM-based sentiment analysis. The DAO module dynamically adjusts task weights based on their relative importance and data characteristics, addressing inter-task difficulty and data imbalance issues. This plug-and-play module can be seamlessly integrated into existing models, enhancing their adaptability to diverse tasks and datasets. Combined with LoRA for efficient fine-tuning, our approach achieves state-of-the-art performance in financial sentiment analysis, significantly improving MSE and accuracy over previous methods.

In the future work, we will explore the use of data-aware classification tasks to enhance the performance of multi-task learning with DAO module under data-limited scenarios.

\appendix

\bibliography{aaai25}

\appendix
\section{Appendix}

\subsection{Dataset}
In this work, the text dataset comes from another study \cite{ding2024eurusd}. Specifically, the dataset is collected from investing.com and forexempire.com, focusing on the EUR/USD exchange rate. The dataset spans from February 6, 2016, to January 19, 2024, encompassing all accessible data on these platforms, resulting in a total of 35,427 records. To address the presence of noise and information irrelevant to the target exchange rate, we use ChatGPT-4.0 and prompt engineering techniques to filter the raw dataset. Further analysis reveals that typically only individual paragraphs or multiple sentences within articles directly relate to the EUR/USD exchange rate, likely catering to readers' diverse interests. To extract the most relevant parts and refine the dataset, we process the text data using ChatGPT-4.0, yielding a final text dataset comprising 23,242 records.

Sentiment polarity annotation in exchange rate texts is particularly complex because such texts are often filled with professional terminology, implied emotions related to market conditions, and subtle variations across industries. Moreover, as exchange rate issues involve two countries, significant positive or negative news about one country can have a substantial impact on financial markets. For example, a large amount of positive news about the United States may strengthen the US dollar, typically leading to a decrease in the EUR/USD exchange rate, while negative news may result in the opposite effect. However, news texts often contain information about both countries simultaneously, making accurate classification of the text complex, with news from the two countries exhibiting a zero-sum effect in sentiment analysis.

To address this challenge, we leverage LLMs to annotate the text with sentiment polarity scores. Large language models have advantages in understanding long text dependencies, complex semantics, and processing high-noise text. For this purpose, we use ChatGPT-4.0 integrated with carefully designed and iteratively refined prompts to annotate the training dataset. To guard against potential and unidentified errors, the model is also required to provide reasons for its assigned sentiment scores. In our prompt engineering, we define the sentiment score range as [-1, 1], with scores close to 1 indicating strong positive sentiment and vice versa. Scores near zero represent neutral sentiment, ensuring comprehensive coverage of the subtle emotional details present in financial narratives.

\subsection{Evaluation Metrics}
For the sentiment score regression task, where the model predicts a continuous sentiment polarity score, we employ the following metrics:

\begin{itemize}
    \item \textbf{Mean Squared Error (MSE)} calculates the average squared differences between the predicted sentiment polarity scores (\(\hat{y}_i\)) and the annotated sentiment polarity scores (\(y_i\)) in the test set. MSE provides a measure of the model's accuracy in predicting the exact sentiment scores. It is defined as:
    \[
    \text{MSE} = \frac{1}{n} \sum_{i=1}^{n} (y_i - \hat{y}_i)^2
    \]
    where \(n\) is the number of texts in the test set.

    \item \textbf{Mean Absolute Error (MAE)} measures the average magnitude of the absolute errors between the predicted and annotated sentiment polarity scores, ignoring their direction. MAE helps in understanding the average error magnitude and is less sensitive to outliers compared to MSE. It is formulated as:
    \[
    \text{MAE} = \frac{1}{n} \sum_{i=1}^{n} \left| y_i - \hat{y}_i \right|
    \]

    \item \textbf{Root Mean Squared Error (RMSE)} is the square root of MSE and provides the error magnitude in the same units as the sentiment polarity scores. RMSE is more interpretable than MSE and is calculated as:
    \[
    \text{RMSE} = \sqrt{\text{MSE}}
    \]

    \item \textbf{R-squared (R\(^2\))} indicates the proportion of variance in the sentiment polarity scores that can be explained by the model's predictions. It provides a measure of how well the model fits the data, with values closer to 1 indicating a better fit. R-squared is defined as:
    \[
    R^2 = 1 - \frac{\sum_{i=1}^{n} (y_i - \hat{y}_i)^2}{\sum_{i=1}^{n} (y_i - \bar{y})^2}
    \]
    where \(\bar{y}\) represents the mean of the annotated sentiment polarity scores in the test set.
\end{itemize}

For the sentiment classification task, where texts are categorized into five sentiment classes (e.g., strong positive, positive, neutral, negative, strong negative), we calculate the following metrics:

\begin{itemize}
    \item \textbf{Accuracy (ACC)}: the ratio of correctly predicted sentiment classes to the total number of predictions. It measures the overall correctness of the model's classifications and is defined as:
    \[
    \text{ACC} = \frac{\sum_{i=1}^{5} TP_i}{\sum_{i=1}^{5} (TP_i + FN_i)}
    \]
    where \( TP_i \) is the number of instances correctly predicted as class \( i \), and \( FN_i \) is the number of instances that actually belong to class \( i \) but are wrongly predicted as other classes.

    \item \textbf{Precision}: the ratio of correctly predicted positive instances to all instances predicted as positive. It measures the model's ability  to avoid false positives and is calculated as:
    \[
    \text{Precision} = \frac{TP_{pos}}{TP_{pos} + FP_{pos}}
    \]
    where \( TP_{pos} \) is the number of positive instances correctly predicted as positive, and \( FP_{pos} \) is the number of non-positive instances wrongly predicted as positive.

    \item \textbf{Recall (or Sensitivity)}: the ratio of correctly predicted positive instances to all actual positive instances. It measures the model's ability to identify all positive instances and is calculated as:
    \[
    \text{Recall} = \frac{TP_{pos}}{TP_{pos} + FN_{pos}}
    \]
    where \( FN_{pos} \) is the number of positive instances wrongly predicted as non-positive. 

    \item \textbf{F1 Score}: the harmonic mean of Precision and Recall, providing a balanced measure of the model's performance, especially when the sentiment classes are imbalanced. It is calculated as:
    \[
    F1 = 2 \times \frac{\text{Precision} \times \text{Recall}}{\text{Precision} + \text{Recall}}
    \]
\end{itemize}

\textbf{Proof of Equivalence Between Weighted Recall and Accuracy in Multi-Class Classification}

In a multi-class classification problem with \( n \) classes, let \( TP_i \), \( FN_i \), and \( \text{Support}_i \) denote the true positives, false negatives, and total instances (support) for class \( i \), respectively. The recall for class \( i \) is given by:
\[
\text{Recall}_i = \frac{TP_i}{TP_i + FN_i} = \frac{TP_i}{\text{Support}_i}
\]
The weighted recall is then calculated as:
\[
\text{Weighted Recall} = \frac{\sum_{i=1}^{n} \text{Recall}_i \times \text{Support}_i}{\sum_{i=1}^{n} \text{Support}_i}
\]
Substituting the expression for \( \text{Recall}_i \):
\[
\text{Weighted Recall} = \frac{\sum_{i=1}^{n} \frac{TP_i}{\text{Support}_i} \times \text{Support}_i}{\sum_{i=1}^{n} \text{Support}_i}
\]
The \( \text{Support}_i \) terms in the numerator and denominator cancel out, leaving:
\[
\text{Weighted Recall} = \frac{\sum_{i=1}^{n} TP_i}{\sum_{i=1}^{n} \text{Support}_i}
\]
This is exactly the formula for overall accuracy:
\[
\text{Accuracy} = \frac{TP_1 + TP_2 + \ldots + TP_n}{\text{Support}_1 + \text{Support}_2 + \ldots + \text{Support}_n}
\]
Therefore, in a multi-class classification setting, weighted recall is mathematically equivalent to accuracy. 

\subsection{Multi-task Loss}

\begin{figure}[!htb]
    \centering
    \includegraphics[width=1\linewidth]{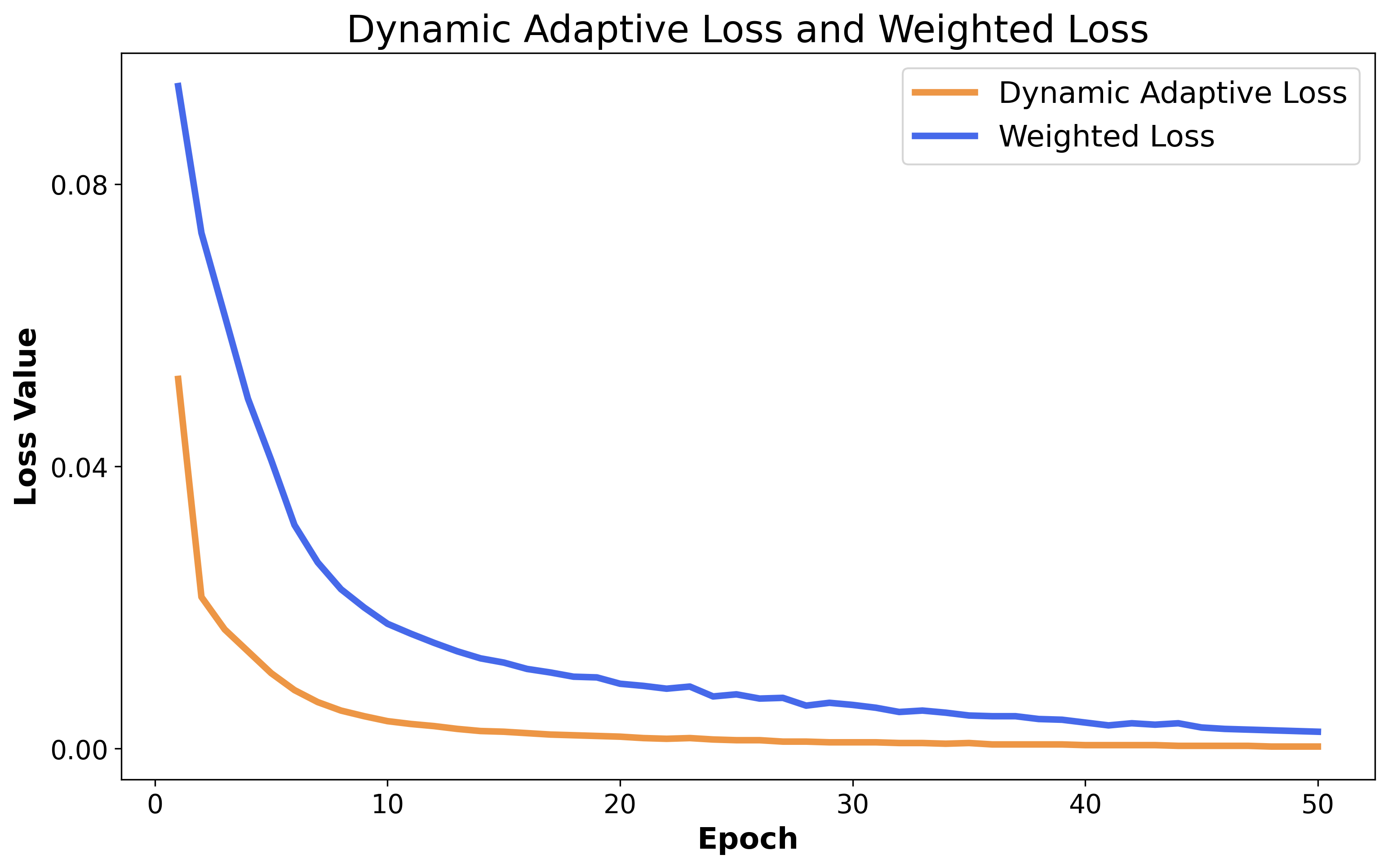}
    \caption{Comparison between constant-weight loss and DAO loss}
    \label{fig:adaptive_loss}
\end{figure}

\begin{figure}[!htb]
    \centering
    \includegraphics[width=1\linewidth]{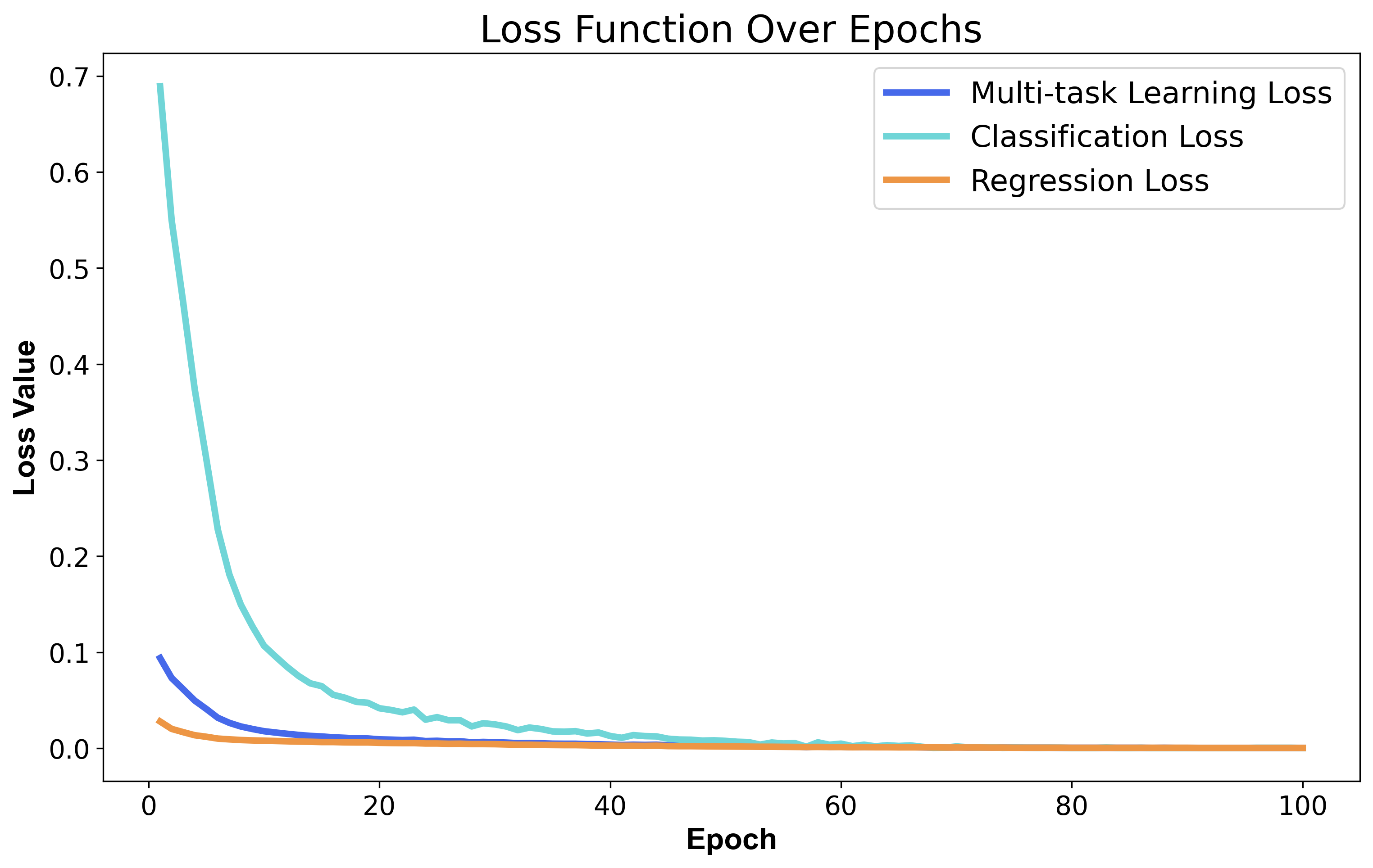}
    \caption{Constant-weight loss}
    \label{fig:adaptive_loss1}
\end{figure}

\begin{figure}[!htb]
    \centering
    \includegraphics[width=1\linewidth]{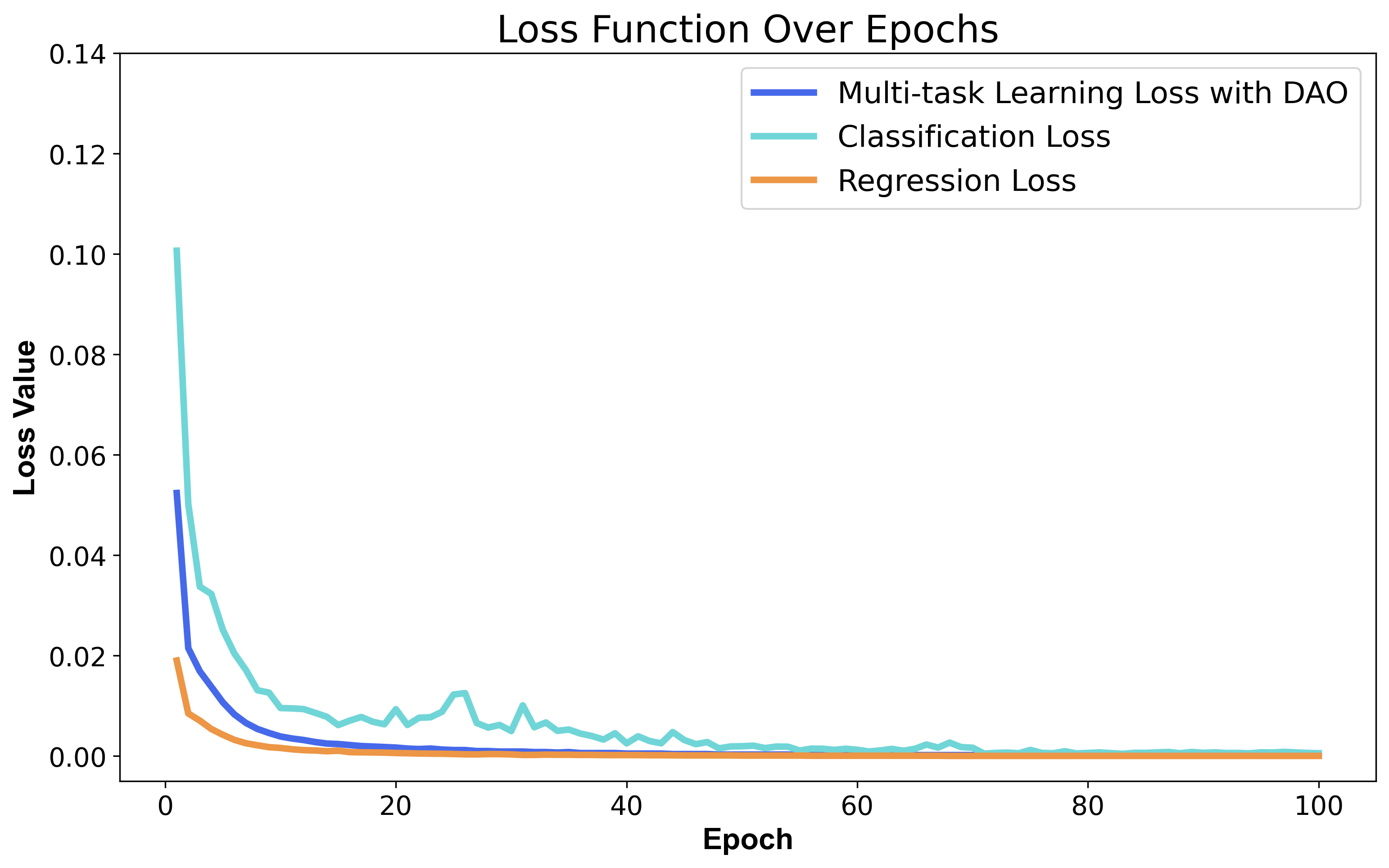}
    \caption{DAO loss}
    \label{fig:adaptive_loss2}
\end{figure}

\subsection{LoRA}
\begin{table*}[!htb]
\centering
\begin{tabular}{C{0.9cm}C{1.3cm}C{1.25cm}C{1.25cm}C{1.35cm}C{1.45cm}C{1.7cm}C{1.7cm}C{1.3cm}C{1.3cm}}
\toprule
Epoch & MSE & MAE & RMSE & R\textsuperscript{2} (\%) & ACC (\%) & Pre. (\%) & Recall (\%) & F1 (\%) & GPU \\
\midrule
20  & 0.022544 & 0.105743 & 0.150147 & 74.68 & 74.83 & 76.37 & 74.83 & 75.39 & 21,780 \\
40  & 0.022930 & 0.107780 & 0.151428 & 74.25 & 75.90 & 76.39 & 75.90 & 76.01 & MiB \\
\cline{10-10}
60  & 0.023744 & 0.107770 & 0.154089 & 73.33 & 75.52 & 75.94 & 75.52 & 75.61 & \textbf{Time} \\
\cline{10-10}
80  & 0.022612 & 0.104911 & 0.150373 & 74.61 & 76.08 & 75.97 & 76.08 & 75.88 & 129,959 \\
100 & 0.022326 & 0.104055 & 0.149420 & 74.93 & 76.11 & 76.17 & 76.11 & 76.05 & seconds \\
\bottomrule
\end{tabular}
\caption{Performance Metrics of Multi-Task Learning with constant Weight.}
\label{table_constant_weight}
\end{table*}

\begin{table*}[!htb]
\centering
\begin{tabular}{C{0.9cm}C{1.3cm}C{1.25cm}C{1.25cm}C{1.35cm}C{1.45cm}C{1.7cm}C{1.7cm}C{1.3cm}C{1.3cm}}
\toprule
Epoch & MSE & MAE & RMSE & R\textsuperscript{2} (\%) & ACC (\%) & Pre. (\%) & Recall (\%) & F1 (\%) & GPU \\
\midrule
20 & 0.018983 & 0.099503 & 0.137889 & 78.65 & 75.60 & 76.99 & 75.60 & 75.99 & 22,336 \\
40 & 0.018668 & 0.097774 & 0.136370 & 79.11 & 76.64 & 77.92 & 76.64 & 76.83 & MiB \\
\cline{10-10}
60 & 0.018841 & 0.094788 & 0.134705 & 79.62 & 76.68 & 78.06 & 76.68 & 76.94 & \textbf{Time} \\
\cline{10-10}
80 & 0.019000 & 0.094113 & 0.133954 & 79.85 & 76.89 & 77.98 & 76.89 & 77.00 & 211,024 \\
100 & 0.018848 & 0.093999 & 0.133965 & 79.85 & 77.06 & 78.28 & 77.06 & 77.37 & seconds \\
\bottomrule
\end{tabular}
\caption{Performance Metrics of Multi-Task Learning with DAO.}
\label{Performance Metrics of Multi-Task Learning with DAO}
\end{table*}

\begin{table*}[!htb]
\centering
\begin{tabular}{C{0.9cm}C{1.3cm}C{1.25cm}C{1.25cm}C{1.35cm}C{1.45cm}C{1.7cm}C{1.7cm}C{1.3cm}C{1.3cm}}
\toprule
Epoch & MSE & MAE & RMSE & R\textsuperscript{2} (\%) & ACC (\%) & Pre. (\%) & Recall (\%) & F1 (\%) & GPU \\
\midrule
20  & 0.020452 & 0.105012 & 0.143009 & 77.03 & 74.18 & 77.08 & 74.18 & 74.99 & 21,872 \\
40  & 0.021576 & 0.104977 & 0.146887 & 75.77 & 73.97 & 76.17 & 73.97 & 74.47 & MiB \\
\cline{10-10}
60  & 0.022072 & 0.105496 & 0.148566 & 75.21 & 74.61 & 76.19 & 74.61 & 74.97 & \textbf{Time} \\
\cline{10-10}
80  & 0.022299 & 0.106003 & 0.149330 & 74.96 & 74.01 & 76.20 & 74.01 & 74.65 & 229,915 \\
100 & 0.022443 & 0.106440 & 0.149811 & 74.80 & 74.23 & 76.46 & 74.23 & 74.93 & seconds \\
\bottomrule
\end{tabular}
\caption{Performance Metrics of Multi-Task Learning with DAO Using LoRA with Rank 8.}
\label{Performance Metrics of Multi-Task Learning with DAO Using LoRA with Rank 8}
\end{table*}

\begin{table*}[!htb]
\centering
\begin{tabular}{C{0.9cm}C{1.3cm}C{1.25cm}C{1.25cm}C{1.35cm}C{1.45cm}C{1.7cm}C{1.7cm}C{1.3cm}C{1.3cm}}
\toprule
Epoch & MSE & MAE & RMSE & R\textsuperscript{2} (\%) & ACC (\%) & Pre. (\%) & Recall (\%) & F1 (\%) & GPU \\
\midrule
20  & 0.023188 & 0.111574 & 0.152275 & 73.96 & 73.06 & 76.55 & 73.06 & 74.32 & 22,248 \\
40  & 0.023726 & 0.110474 & 0.154032 & 73.35 & 73.67 & 75.84 & 73.67 & 74.45 & MiB \\
\cline{10-10}
60  & 0.023067 & 0.108082 & 0.151877 & 74.10 & 73.41 & 75.51 & 73.41 & 73.96 & \textbf{Time} \\
\cline{10-10}
80  & 0.022849 & 0.108018 & 0.151159 & 74.34 & 73.58 & 75.48 & 73.58 & 74.14 & 234,485 \\
100 & 0.022653 & 0.107673 & 0.150508 & 74.56 & 74.14 & 75.80 & 74.14 & 74.63 & seconds \\
\bottomrule
\end{tabular}
\caption{Performance Metrics of Multi-Task Learning with DAO Using LoRA with Rank 16.}
\label{Performance Metrics of Multi-Task Learning with DAO Using LoRA with Rank 16}
\end{table*}

\begin{table*}[!htb]
\centering
\begin{tabular}{C{0.9cm}C{1.3cm}C{1.25cm}C{1.25cm}C{1.35cm}C{1.45cm}C{1.7cm}C{1.7cm}C{1.3cm}C{1.3cm}}
\toprule
Epoch & MSE & MAE & RMSE & R\textsuperscript{2} (\%) & ACC (\%) & Pre. (\%) & Recall (\%) & F1 (\%) & GPU \\
\midrule
20  & 0.022356 & 0.107567 & 0.149518 & 74.89 & 73.36 & 76.56 & 73.36 & 74.38 & 22,248 \\
40  & 0.022449 & 0.107912 & 0.149831 & 74.79 & 74.61 & 77.02 & 74.61 & 75.49 & MiB \\
\cline{10-10}
60  & 0.021086 & 0.103995 & 0.145209 & 76.32 & 74.01 & 76.20 & 74.01 & 74.72 & \textbf{Time} \\
\cline{10-10}
80  & 0.020559 & 0.102930 & 0.143382 & 76.91 & 74.61 & 76.27 & 74.61 & 75.08 & 234,405 \\
100 & 0.020758 & 0.103383 & 0.144078 & 76.69 & 74.66 & 76.44 & 74.66 & 75.23 & seconds \\
\bottomrule
\end{tabular}
\caption{Performance Metrics of Multi-Task Learning with DAO Using LoRA with Rank 32.}
\label{Performance Metrics of Multi-Task Learning with DAO Using LoRA with Rank 32}
\end{table*}

\begin{table*}[!htb]
\centering
\begin{tabular}{C{0.9cm}C{1.3cm}C{1.25cm}C{1.25cm}C{1.35cm}C{1.45cm}C{1.7cm}C{1.7cm}C{1.3cm}C{1.3cm}}
\toprule
Epoch & MSE & MAE & RMSE & R\textsuperscript{2} (\%) & ACC (\%) & Pre. (\%) & Recall (\%) & F1 (\%) & GPU \\
\midrule
20  & 0.021734 & 0.106654 & 0.147425 & 75.59 & 74.27 & 76.28 & 74.27 & 74.97 & 22,248 \\
40  & 0.020599 & 0.102518 & 0.143523 & 76.87 & 74.40 & 75.99 & 74.40 & 74.69 & MiB \\
\cline{10-10}
60  & 0.020051 & 0.101035 & 0.141602 & 77.48 & 75.04 & 76.95 & 75.04 & 75.62 & \textbf{Time} \\
\cline{10-10}
80  & 0.019859 & 0.100704 & 0.140922 & 77.70 & 74.70 & 76.45 & 74.70 & 75.27 & 234,230 \\
100 & 0.019768 & 0.100669 & 0.140598 & 77.80 & 74.87 & 76.83 & 74.87 & 75.51 & seconds \\
\bottomrule
\end{tabular}
\caption{Performance Metrics of Multi-Task Learning with DAO Using LoRA with Rank 64.}
\label{Performance Metrics of Multi-Task Learning with DAO Using LoRA with Rank 64}
\end{table*}

\begin{table*}[!htb]
\centering
\begin{tabular}{C{0.9cm}C{1.3cm}C{1.25cm}C{1.25cm}C{1.35cm}C{1.45cm}C{1.7cm}C{1.7cm}C{1.3cm}C{1.3cm}}
\toprule
Epoch & MSE & MAE & RMSE & R\textsuperscript{2} (\%) & ACC (\%) & Pre. (\%) & Recall (\%) & F1 (\%) & GPU \\
\midrule
20  & 0.020855 & 0.103128 & 0.144412 & 76.58 & 75.13 & 76.73 & 75.13 & 75.56 & 22,852 \\
40  & 0.020442 & 0.102753 & 0.142975 & 77.04 & 75.13 & 77.25 & 75.13 & 75.80 & MiB \\
\cline{10-10}
60  & 0.019320 & 0.099492 & 0.138995 & 78.30 & 76.55 & 78.10 & 76.55 & 76.93 & \textbf{Time} \\
\cline{10-10}
80  & 0.019026 & 0.098734 & 0.137935 & 78.63 & 76.25 & 77.93 & 76.25 & 76.65 & 247,452 \\
100 & 0.019057 & 0.098797 & 0.138049 & 78.60 & 76.33 & 78.02 & 76.33 & 76.76 & seconds \\
\bottomrule
\end{tabular}
\caption{Performance Metrics of Multi-Task Learning with DAO Using LoRA with Rank 128.}
\label{Performance Metrics of Multi-Task Learning with DAO Using LoRA with Rank 128}
\end{table*}

\begin{table*}[!htb]
\centering
\begin{tabular}{C{0.9cm}C{1.3cm}C{1.25cm}C{1.25cm}C{1.35cm}C{1.45cm}C{1.7cm}C{1.7cm}C{1.3cm}C{1.3cm}}
\toprule
Epoch & MSE & MAE & RMSE & R\textsuperscript{2} (\%) & ACC (\%) & Pre. (\%) & Recall (\%) & F1 (\%) & GPU \\
\midrule
20  & 0.020905 & 0.103332 & 0.144585 & 76.52 & 75.13 & 77.44 & 75.13 & 75.78 & 24,224 \\
40  & 0.019155 & 0.098854 & 0.138400 & 78.49 & 76.72 & 78.46 & 76.72 & 77.00 & MiB \\
\cline{10-10}
60  & 0.019064 & 0.098707 & 0.138074 & 78.59 & 76.42 & 78.39 & 76.42 & 76.87 & \textbf{Time} \\
\cline{10-10}
80  & 0.018804 & 0.098018 & 0.137127 & 78.88 & 75.82 & 77.84 & 75.82 & 76.25 & 266,674 \\
100 & 0.018747 & 0.097745 & 0.136919 & 78.95 & 76.03 & 78.03 & 76.03 & 76.47 & seconds \\
\bottomrule
\end{tabular}
\caption{Performance Metrics of Multi-Task Learning with DAO Using LoRA with Rank 256.}
\label{Performance Metrics of Multi-Task Learning with DAO Using LoRA with Rank 256}
\end{table*}

\begin{table*}[!htb]
\centering
\begin{tabular}{C{0.9cm}C{1.3cm}C{1.25cm}C{1.25cm}C{1.35cm}C{1.45cm}C{1.7cm}C{1.7cm}C{1.3cm}C{1.3cm}}
\toprule
Epoch & MSE & MAE & RMSE & R\textsuperscript{2} (\%) & ACC (\%) & Pre. (\%) & Recall (\%) & F1 (\%) & GPU \\
\midrule
20  & 0.020939 & 0.103973 & 0.144705 & 76.48 & 75.17 & 76.82 & 75.17 & 75.76 & 19,550 \\
40  & 0.019411 & 0.100529 & 0.139324 & 78.20 & 75.09 & 76.94 & 75.09 & 75.52 & MiB \\
\cline{10-10}
60  & 0.018673 & 0.097442 & 0.136649 & 79.03 & 75.73 & 77.52 & 75.73 & 76.14 & \textbf{Time} \\
\cline{10-10}
80  & 0.018651 & 0.097285 & 0.136570 & 79.05 & 75.39 & 77.03 & 75.39 & 75.69 & 286,793 \\
100 & 0.018610 & 0.097076 & 0.136419 & 79.10 & 75.47 & 77.04 & 75.47 & 75.68 & seconds \\
\bottomrule
\end{tabular}
\caption{Performance Metrics of Multi-Task Learning with DAO Using LoRA with Rank 384.}
\label{Performance Metrics of Multi-Task Learning with DAO Using LoRA with Rank 384}
\end{table*}

\begin{table*}[!htb]
\centering
\begin{tabular}{C{0.9cm}C{1.3cm}C{1.25cm}C{1.25cm}C{1.35cm}C{1.45cm}C{1.7cm}C{1.7cm}C{1.3cm}C{1.3cm}}
\toprule
Epoch & MSE & MAE & RMSE & R\textsuperscript{2} (\%) & ACC (\%) & Pre. (\%) & Recall (\%) & F1 (\%) & GPU \\
\midrule
20  & 0.020835 & 0.103436 & 0.144345 & 76.60 & 75.00 & 76.68 & 75.00 & 75.66 & 22,336 \\
40  & 0.018324 & 0.096110 & 0.135365 & 79.42 & 76.85 & 78.27 & 76.85 & 76.97 & MiB \\
\cline{10-10}
60  & 0.018376 & 0.095549 & 0.135560 & 79.36 & 76.12 & 77.74 & 76.12 & 76.31 & \textbf{Time} \\
\cline{10-10}
80  & 0.018580 & 0.095503 & 0.136307 & 79.13 & 76.76 & 77.67 & 76.76 & 76.65 & 324,013 \\
100 & 0.018571 & 0.095562 & 0.136276 & 79.14 & 76.38 & 77.40 & 76.38 & 76.35 & seconds \\
\bottomrule
\end{tabular}
\caption{Performance Metrics of Multi-Task Learning with DAO Using LoRA with Rank 512.}
\label{Performance Metrics of Multi-Task Learning with DAO Using LoRA with Rank 512}
\end{table*}

\subsection{Multi-task Learning with constant Weights}

\begin{table*}[!htb]
\centering
\begin{tabular}{C{0.9cm}C{1.3cm}C{1.25cm}C{1.25cm}C{1.35cm}C{1.45cm}C{1.7cm}C{1.7cm}C{1.3cm}C{1.3cm}}
\toprule
Epoch & MSE & MAE & RMSE & R\textsuperscript{2} (\%) & ACC (\%) & Pre. (\%) & Recall (\%) & F1 (\%) & GPU \\
\midrule
20  & 0.020224 & 0.101586 & 0.142211 & 77.29 & 77.15 & 76.25 & 77.15 & 76.39 & 21,780 \\
40  & 0.021843 & 0.103883 & 0.147792 & 75.47 & 76.72 & 76.12 & 76.72 & 76.31 & MiB \\
\cline{10-10}
60  & 0.021938 & 0.103734 & 0.148115 & 75.36 & 76.29 & 75.92 & 76.29 & 75.83 & \textbf{Time} \\
\cline{10-10}
80  & 0.021660 & 0.101345 & 0.147174 & 75.67 & 76.85 & 76.19 & 76.85 & 76.33 & 129,943 \\
100 & 0.021260 & 0.100958 & 0.145807 & 76.12 & 76.81 & 76.16 & 76.81 & 76.32 & seconds \\
\bottomrule
\end{tabular}
\caption{Performance Metrics of Multi-Task Learning with constant Weight 0.05.}
\label{Performance Metrics of Multi-Task Learning with constant Weight 0.05}
\end{table*}

\begin{table*}[!htb]
\centering
\begin{tabular}{C{0.9cm}C{1.3cm}C{1.25cm}C{1.25cm}C{1.35cm}C{1.45cm}C{1.7cm}C{1.7cm}C{1.3cm}C{1.3cm}}
\toprule
Epoch & MSE & MAE & RMSE & R\textsuperscript{2} (\%) & ACC (\%) & Pre. (\%) & Recall (\%) & F1 (\%) & GPU \\
\midrule
20  & 0.022544 & 0.105743 & 0.150147 & 74.68 & 74.83 & 76.37 & 74.83 & 75.39 & 21,780 \\
40  & 0.022930 & 0.107780 & 0.151428 & 74.25 & 75.90 & 76.39 & 75.90 & 76.01 & MiB \\
\cline{10-10}
60  & 0.023744 & 0.107770 & 0.154089 & 73.33 & 75.52 & 75.94 & 75.52 & 75.61 & \textbf{Time} \\
\cline{10-10}
80  & 0.022612 & 0.104911 & 0.150373 & 74.61 & 76.08 & 75.97 & 76.08 & 75.88 & 129,959 \\
100 & 0.022326 & 0.104055 & 0.149420 & 74.93 & 76.11 & 76.17 & 76.11 & 76.05 & seconds \\
\bottomrule
\end{tabular}
\caption{Performance Metrics of Multi-Task Learning with constant Weight 0.10.}
\label{table_constant_weight_0.1}
\end{table*}

\begin{table*}[!htb]
\centering
\begin{tabular}{C{0.9cm}C{1.3cm}C{1.25cm}C{1.25cm}C{1.35cm}C{1.45cm}C{1.7cm}C{1.7cm}C{1.3cm}C{1.3cm}}
\toprule
Epoch & MSE & MAE & RMSE & R\textsuperscript{2} (\%) & ACC (\%) & Pre. (\%) & Recall (\%) & F1 (\%) & GPU \\
\midrule
20  & 0.021989 & 0.104959 & 0.148285 & 75.31 & 76.08 & 75.31 & 76.08 & 75.53 & 21,780 \\
40  & 0.024872 & 0.111896 & 0.157707 & 72.07 & 73.84 & 73.88 & 73.84 & 73.66 & MiB \\
\cline{10-10}
60  & 0.021891 & 0.105077 & 0.147954 & 75.42 & 76.38 & 75.25 & 76.38 & 75.44 & \textbf{Time} \\
\cline{10-10}
80  & 0.022689 & 0.105070 & 0.150628 & 74.52 & 76.72 & 75.75 & 76.72 & 76.06 & 130,013 \\
100 & 0.022417 & 0.104278 & 0.149722 & 74.83 & 76.46 & 75.67 & 76.46 & 75.93 & seconds \\
\bottomrule
\end{tabular}
\caption{Performance Metrics of Multi-Task Learning with constant Weight 0.15.}
\label{Performance Metrics of Multi-Task Learning with constant Weight 0.15}
\end{table*}

\begin{table*}[!htb]
\centering
\begin{tabular}{C{0.9cm}C{1.3cm}C{1.25cm}C{1.25cm}C{1.35cm}C{1.45cm}C{1.7cm}C{1.7cm}C{1.3cm}C{1.3cm}}
\toprule
Epoch & MSE & MAE & RMSE & R\textsuperscript{2} (\%) & ACC (\%) & Pre. (\%) & Recall (\%) & F1 (\%) & GPU \\
\midrule
20  & 0.023922 & 0.108378 & 0.154667 & 73.13 & 75.56 & 75.28 & 75.56 & 75.28 & 21,780 \\
40  & 0.025132 & 0.110748 & 0.158530 & 71.78 & 74.83 & 74.88 & 74.83 & 74.84 & MiB \\
\cline{10-10}
60  & 0.024118 & 0.107101 & 0.155301 & 72.91 & 75.86 & 75.01 & 75.86 & 75.33 & \textbf{Time} \\
\cline{10-10}
80  & 0.024253 & 0.107076 & 0.155735 & 72.76 & 75.77 & 75.21 & 75.77 & 75.40 & 129,972 \\
100 & 0.023748 & 0.106396 & 0.154105 & 73.33 & 75.77 & 74.99 & 75.77 & 75.30 & seconds \\
\bottomrule
\end{tabular}
\caption{Performance Metrics of Multi-Task Learning with constant Weight 0.20.}
\label{Performance Metrics of Multi-Task Learning with constant Weight 0.20}
\end{table*}

\begin{table*}[!htb]
\centering
\begin{tabular}{C{0.9cm}C{1.3cm}C{1.25cm}C{1.25cm}C{1.35cm}C{1.45cm}C{1.7cm}C{1.7cm}C{1.3cm}C{1.3cm}}
\toprule
Epoch & MSE & MAE & RMSE & R\textsuperscript{2} (\%) & ACC (\%) & Pre. (\%) & Recall (\%) & F1 (\%) & GPU \\
\midrule
20  & 0.021961 & 0.107441 & 0.148191 & 75.34 & 75.00 & 75.07 & 75.00 & 74.93 & 21,780 \\
40  & 0.023688 & 0.107082 & 0.153909 & 73.40 & 75.77 & 76.07 & 75.77 & 75.90 & MiB \\
\cline{10-10}
60  & 0.023203 & 0.106424 & 0.152324 & 73.94 & 76.76 & 76.26 & 76.76 & 76.32 & \textbf{Time} \\
\cline{10-10}
80  & 0.022665 & 0.105023 & 0.150549 & 74.55 & 77.71 & 77.06 & 77.71 & 77.13 & 129,889 \\
100 & 0.022714 & 0.104991 & 0.150712 & 74.49 & 76.98 & 76.44 & 76.98 & 76.60 & seconds \\
\bottomrule
\end{tabular}
\caption{Performance Metrics of Multi-Task Learning with constant Weight 0.25.}
\label{Performance Metrics of Multi-Task Learning with constant Weight 0.25}
\end{table*}

\begin{table*}[!htbp]
\centering
\begin{tabular}{C{0.9cm}C{1.3cm}C{1.25cm}C{1.25cm}C{1.35cm}C{1.45cm}C{1.7cm}C{1.7cm}C{1.3cm}C{1.3cm}}
\toprule
Epoch & MSE & MAE & RMSE & R\textsuperscript{2} (\%) & ACC (\%) & Pre. (\%) & Recall (\%) & F1 (\%) & GPU \\
\midrule
20  & 0.024470 & 0.110197 & 0.156430 & 72.52 & 74.35 & 74.05 & 74.35 & 74.03 & 21,780 \\
40  & 0.023319 & 0.107901 & 0.152704 & 73.81 & 75.56 & 75.12 & 75.56 & 75.29 & MiB \\
\cline{10-10}
60  & 0.023074 & 0.106771 & 0.151901 & 74.09 & 75.82 & 75.94 & 75.82 & 75.63 & \textbf{Time} \\
\cline{10-10}
80  & 0.023687 & 0.107406 & 0.153907 & 73.40 & 76.16 & 75.32 & 76.16 & 75.61 & 130,001 \\
100 & 0.023137 & 0.105845 & 0.152110 & 74.02 & 76.46 & 75.77 & 76.46 & 76.03 & seconds \\
\bottomrule
\end{tabular}
\caption{Performance Metrics of Multi-Task Learning with constant Weight 0.30.}
\label{Performance Metrics of Multi-Task Learning with constant Weight 0.30}
\end{table*}

\begin{table*}[!htbp]
\centering
\begin{tabular}{@{}C{0.8cm}C{0.8cm}C{1.2cm}C{1.2cm}C{1.2cm}C{1.0cm}C{1.0cm}C{1.0cm}C{1.0cm}C{1.0cm}C{1.5cm}C{1.45cm}@{}}
\toprule
Rank & Epoch & MSE & MAE & RMSE & R\textsuperscript{2} & ACC & Pre. & Recall & F1 & Time & Usage \\
\midrule
128 & 85 & 0.018977 & 0.098725 & 0.137757 & 78.69 & 76.51 & 78.15 & 76.51 & 76.89 & 210,333 s & 99.67\% \\
256 & 64 & 0.018838 & 0.098094 & 0.137250 & 78.84 & 76.16 & 77.99 & 76.16 & 76.50 & 170,671 s & 80.88\% \\
256 & 79 & 0.018665 & 0.097445 & 0.136619 & 79.04 & 75.82 & 77.76 & 75.82 & 76.25 & 210,672 s & 99.83\% \\
384 & 39 & 0.018434 & 0.096930 & 0.135773 & 79.30 & 75.69 & 77.80 & 75.69 & 76.20 & 111,849 s & 53.00\% \\
512 & 23 & 0.018730 & 0.098810 & 0.136857 & 78.97 & 76.38 & 77.58 & 76.38 & 76.58 & 77,763 s & 36.85\% \\
512 & 42 & 0.018154 & 0.095840 & 0.134738 & 79.61 & 77.15 & 78.32 & 77.15 & 77.02 & 139,326 s & 66.02\% \\
\bottomrule
\end{tabular}
\caption{Performance Metrics of LORA}
\label{tab:performance_metrics_B2}
\end{table*}

\subsection{Related Work}

\noindent{\textbf{Sentiment Analysis}} Sentiment analysis is widely used across various domains. In the financial sector, \citeauthor{li2019text} (\citeyear{li2019text}) and \citeauthor{correia2022deep} (\citeyear{correia2022deep}) employ these methods to improve forecasting accuracy for crude oil prices and stock market movements, respectively. Extending this approach, \citeauthor{qian2022understanding} (\citeyear{qian2022understanding}) examines NFT-related tweets to correlate public sentiment with market trends. Beyond finance, \citeauthor{wen2024text} (\citeyear{wen2024text}) utilizes text mining on online reviews to assess product competitiveness, while \citeauthor{garner2022utilizing} (\citeyear{garner2022utilizing}) applies similar techniques to analyze factors influencing consumer happiness in travel experiences. These studies collectively underscore the versatility and effectiveness of text analysis methods in extracting valuable insights from unstructured data across diverse fields.

\noindent{\textbf{Lexicon-based Methods}} Lexicon-based methods for sentiment analysis, while effective across various domains, face limitations due to their reliance on predefined sentiment dictionaries. Studies demonstrate their application in diverse fields. \citeauthor{barik2024analysis} (\citeyear{barik2024analysis}) and \citeauthor{liu2020consumers} (\citeyear{liu2020consumers}) develop models for multi-domain sentiment analysis and online pharmacy reviews, respectively. During the COVID-19 pandemic, \citeauthor{khan2021us} (\citeyear{khan2021us}), \citeauthor{marcec2022using} (\citeyear{marcec2022using}), and \citeauthor{samaras2023sentiment} (\citeyear{samaras2023sentiment}) apply these methods to analyze public sentiment through social media data. These studies collectively highlight the versatility of lexicon-based approaches while acknowledging potential constraints in lexicon coverage and quality.

\noindent{\textbf{Traditional Methods}} Recent studies explore various machine learning approaches for sentiment analysis across diverse domains. Traditional algorithms like SVM, Random Forest, and Naïve Bayes are applied by \citeauthor{bengesi2023machine} (\citeyear{bengesi2023machine}) for analyzing public sentiment on disease outbreaks, \citeauthor{ranibaran2021analyzing} (\citeyear{ranibaran2021analyzing}) for stock price prediction, and \citeauthor{asif2020sentiment} (\citeyear{asif2020sentiment}) for multilingual extremism text classification. \citeauthor{naresh2021efficient} (\citeyear{naresh2021efficient}) proposes a hybrid algorithm for Twitter sentiment analysis, while \citeauthor{gopi2023classification} (\citeyear{gopi2023classification}) and \citeauthor{budhi2021using} (\citeyear{budhi2021using}) focus on movie reviews and online ratings respectively. Advanced neural networks, including the LSIBA-ENN by \citeauthor{zhao2021machine} (\citeyear{zhao2021machine}), CNNs and LSTMs by \citeauthor{meena2022categorizing} (\citeyear{meena2022categorizing}), and a two-state GRU model by \citeauthor{zulqarnain2024efficient} (\citeyear{zulqarnain2024efficient}), show promising results in various sentiment classification tasks. \citeauthor{alsayat2022improving} (\citeyear{alsayat2022improving}) proposes an ensemble deep learning model combining FastText and LSTM. Transformer-based methods, like DICET introduced by \citeauthor{naseem2020transformer} (\citeyear{naseem2020transformer}), further advance Twitter sentiment analysis. \citeauthor{aslam2022sentiment} (\citeyear{aslam2022sentiment}) develops an LSTM-GRU ensemble for cryptocurrency-related sentiment analysis. While these studies demonstrate the effectiveness of various techniques across different applications, they often lack comprehensive comparisons and discussions of potential limitations such as overfitting, computational complexity, and generalizability.

\end{document}